%% file: paper.tex
\documentclass[sigconf,nonacm]{acmart}

\usepackage{wrapfig}
\usepackage{threeparttable}
\usepackage{multirow}
\usepackage{tikz}
\usepackage{pgfplotstable}
\usepackage{xcolor}
\usepackage{subcaption}
\usepackage{enumitem}
\usepackage{float}
\usepackage{bm}
\usepackage{stfloats}
\usepackage{pgfplots}
\usepackage[ruled,vlined,linesnumbered]{algorithm2e} 
\usepackage{xspace}
\usepackage{pifont}

\usepackage{hyperref}
\hypersetup{
  colorlinks,breaklinks,
            urlcolor=[RGB]{0,0.5,0.5},
            linkcolor=[RGB]{0,0.5,0.5}
}

\pgfplotsset{compat=1.18}
    \def\addlegendimage{\csname pgfplots@addlegendimage\endcsname}

\AtBeginDocument{%
  \providecommand\BibTeX{{%
    \normalfont B\kern-0.5em{\scshape i\kern-0.25em b}\kern-0.8em\TeX}}}

\setcopyright{acmlicensed}

\acmConference[ACM SIGKDD]{Make sure to enter the correct
  conference title from your rights confirmation emai}{Aug 3--7,
  2025}{Toronto, Canada}
  
\input{dfn.tex}

\begin{document}


\title[Uncertainty-aware Human Mobility  Modeling and Anomaly Detection]{Modeling Aleatoric and Epistemic Uncertainty for  Human Mobility  and Anomaly Detection}
\author{Haomin Wen}
\authornote{Equal contribution}
\affiliation{%
  \institution{Carnegie Mellon University}
   \country{Pittsburgh,USA}
  }
\email{haominwe@andrew.cmu.edu}
\author{Shurui Cao}
\authornotemark[1]
\affiliation{%
  \institution{Carnegie Mellon University}
  \country{Pittsburgh, USA}
 }
\email{shuruic@andrew.cmu.edu}

\author{Zeeshan Rasheed}
\affiliation{%
  \institution{Novateur Research Solutions}
  \country{Ashburn, USA}
  }
\email{zrasheed@novateur.ai}

\author{Khurram Hassan Shafique}
\affiliation{%
  \institution{Novateur Research Solutions}
  \country{Ashburn, USA}
  }
\email{kshafique@novateur.ai}

\author{Leman Akoglu}
\affiliation{%
  \institution{Carnegie Mellon University}
  \country{Pittsburgh, USA}
  }
\email{lakoglu@andrew.cmu.edu}




\renewcommand{\shortauthors}{Haomin Wen et al.}

\begin{abstract}
\input{00abstract.tex}

\end{abstract}


\maketitle

\section{Introduction}
\label{sec:intro}

\input{01intro}

\section{Problem Formulation}
\label{sec:prelim}

\input{02prelim}

\section{Proposed Model}
\label{sec:blocks}

\input{03method}
\section{Experiments}
\label{sec:experiments}
\input{04experiments}
\section{Related Work}
 \vspace{-0.05in}
\label{sec:related}

\input{05relatedwork}

\vspace{-0.05in}
\section{Conclusion}
\label{sec:conclusion}
\input{06conclusion}

\clearpage



\bibliographystyle{ACM-Reference-Format}
\bibliography{00refs}

\clearpage
\appendix
\input{07appendix.tex}

\end{document}

%% file: dfn.tex
\newcommand{\red}[1]{\textcolor{red}{#1}}

\newcommand{\method}{{\sc USTAD}\xspace}
\newcommand{\methodwrej}{{\sc USTAD\_Rej}\xspace}
\newcommand{\trial}{{\small{\textsf{SimLA}}}\xspace}
\newcommand{\trialthree}{{\small{\textsf{SimJordan}}}\xspace}

\newcommand{\NUMOSIM}{{\small{\textsf{NUMOSIM}}}\xspace}
\newcommand{\ssmlp}{{Nov-Online}\xspace}

\newcommand{\R}{\mathbb{R}}
\newcommand{\e}{\bm{e}}
\newcommand{\E}{\bm{E}}

\newcommand{\F}{\mathcal{F}}
\newcommand{\M}{\mathcal{M}}

\makeatletter
\newcommand\footnoteref[1]{\protected@xdef\@thefnmark{\ref{#1}}\@footnotemark}
\makeatother

\newcommand{\cbit}{\begin{compactitem}}
\newcommand{\ceit}{\end{compactitem}}
\newcommand{\cben}{\begin{compactenum}}
\newcommand{\ceen}{\end{compactenum}}

\newcommand{\beq}{\begin{equation}}
	\newcommand{\eeq}{\end{equation}}

\newcommand{\bit}{\begin{itemize}}
	\newcommand{\eit}{\end{itemize}}
\newcommand{\ben}{\begin{enumerate}}
	\newcommand{\een}{\end{enumerate}}

\newcounter{x}

\definecolor{celadon}{rgb}{0.67, 0.88, 0.69}
\definecolor{carolinablue}{rgb}{0.6, 0.73, 0.89}

\definecolor{aliceblue}{rgb}{0.867, 0.917, 0.964}
\definecolor{aliceyellow}{rgb}{0.999, 0.945, 0.796}
\definecolor{alicegray}{rgb}{0.844, 0.867, 0.898}


%% file: 00abstract.tex
Given the temporal GPS coordinates from a large set of human agents, how can we model their mobility behavior toward effective anomaly (e.g. bad-actor or malicious behavior) detection without any labeled data?
Human mobility and trajectory modeling have been extensively studied, showcasing varying abilities to manage complex inputs and balance performance-efficiency trade-offs. In this work, we formulate anomaly detection in complex human behavior by modeling raw GPS data as a sequence of stay-point events, each characterized by spatio-temporal features, along with trips (i.e. commute) between the stay-points. 
Our problem formulation allows us to
leverage modern sequence models 
for unsupervised training and anomaly detection. Notably, we equip our proposed model \method (for Uncertainty-aware Spatio-Temporal Anomaly Detection) with aleatoric (i.e. data) uncertainty estimation to account for inherent stochasticity in certain individuals' behavior, as well as epistemic (i.e. model) uncertainty to handle data sparsity under a large variety of human behaviors.
Together, aleatoric and epistemic uncertainties unlock a robust loss function as well as \sloppy{uncertainty-aware decision-making in anomaly scoring.} Extensive experiments shows that \method improves anomaly detection AUCROC by 3\%-15\%   over baselines in industry-scale data.

%% file: 01intro.tex
\par Human mobility modeling and anomaly detection constitute crucial tasks for diverse applications \cite{barbosa2018human}, from security and surveillance \cite{adey2004surveillance,palm2013rights} to health monitoring \cite{meloni2011modeling,tizzoni2014use}. Effectively identifying anomalies from human GPS data helps uncover critical findings such as disease outbreaks \cite{outliershealth24} 
and security threats \cite{meloni2011modeling,barbosa2018human,numosim}.

However, modeling human activity and detecting anomalies in complex behavior is far from trivial and pose three fundamental challenges: 1) \textbf{Intricate Spatiotemporal Dependencies:} Human mobility data exhibit non-trivial \textit{intra}-event correlations (e.g. the time of day influencing the type of location visited) as well as \textit{inter}-event dependencies (e.g. lunch at a restaurant followed by a return to the office). 2) \textbf{Uncertainty Modeling:} Human behavior is inherently uncertain, characterized by: $i$)  \textit{Aleatoric (data) uncertainty}, which arises from the stochastic nature of individual behavior (e.g., freelancers and tourists engaging in irregular leisure activities, unlike office workers with consistent daily routines), and $ii$) \textit {Epistemic (model) uncertainty}, which stems from data sparsity and limitations in model knowledge. 3) \textbf{Uncertainty-based Anomaly Scoring:~} Even when the aforementioned uncertainties are quantified and behavioral patterns are learned, it remains unclear how to leverage different types of uncertainties, along with deviations from the learned patterns, toward effective anomaly detection.

As shown in Table~\ref{tab:method_compare}, while previous efforts in human mobility and anomaly detection have made significant strides, none of them addresses all the three challenges underscored above. Specifically, human mobility modeling methods \cite{2021CTLE,2023LightPath,DeepMove,want2024trans,MobTCast} focus on capturing spatiotemporal dependencies, but most of them tend to overlook the intrinsic uncertainty. Meanwhile, though uncertainty 
estimation has been extensively studied in probability and statistics \cite{chatfield1995model,smith2024uncertainty}, with increasing interest in uncertainty modeling for deep models \cite{gawlikowski2023survey,chanestimating,kendall2016,Huang2018semantic,sensoy2018evidential,ye2024uncertaintyregularized},
there is still a shortage of work that study and explicitly leverage the aleatoric and epistemic uncertainties for sequence modeling. As for anomaly detection, prior work on trajectory anomaly detection  \cite{IBAT2011, ATROM, song2018anomalous, onlineGMVSAE2020, li2024difftad}  primarily target vehicle movement, which is uniformly sampled and lacks the semantic complexity of human activities. Time series anomaly detection \cite{2024DualTF, Feng2024HUE, zamanzadeh2024deep, wang2024revisiting}, on the other hand, emphasizes time or frequency-domain analysis but struggle to handle discrete events that characterize human activities, as well as mixed-type data with both numerical and categorical features.

To bridge this gap, we present \method, a framework that explicitly quantifies uncertainty for event sequence modeling, analyzes what these uncertainty measures reveal about human behavior, and effectively incorporates them into uncertainty-aware anomaly detection. 
We begin by formulating the problem as an unsupervised learning task, where stay events and trips between them are extracted from raw GPS data, enriched with abundant semantic features. 
\method then employs a ``dual'' Transformer architecture \cite{truong2023poet} with both feature-level and event-level attention, thereby capturing spatio-temporal dependencies. Notably, our model estimates both aleatoric (data) and epistemic (model) uncertainties within a single architecture, providing a comprehensive representation of uncertainty in human mobility. 
By incorporating these uncertainty estimates, \method enables both robust training and reliable inference. Unlike forecasting-based anomaly detection that relies solely on prediction errors, \method integrates uncertainty into its anomaly scoring mechanism, providing a more nuanced and accurate approach to anomaly detection in human behavior. The following summarizes the main contributions of this work.

\input{tables/tab_related}

\begin{itemize}[leftmargin=*]
\itemsep0em
    \item  \textbf{Problem Formulation:} We frame human mobility anomaly detection as an uncertainty-aware sequence modeling problem, transforming time-regular, continuous GPS readings (of the form $\langle x,y,t\rangle$; i.e. coordinates over time) into irregular, discrete stay-point events over time with rich spatio-temporal features that include embeddings of the trips between stay-points. This step resembles a form of raw input ``tokenization'' into a higher-level representation that captures mobility patterns more meaningfully.



    \item \textbf{Uncertainty-aware Dependency Modeling:} 
    We introduce \method, a \textit{dual} Transformer architecture designed to be \textit{uncertainty-aware};  
    which is equipped with ($i$) both feature-level and event-level attention---for capturing dependencies among features as well as across events, respectively; and ($ii$) both data and model uncertainty estimation---toward accounting for inherent stochasticity in human behavior as well as data scarcity, respectively.


    \item \textbf{Uncertainty-aware Anomaly Scoring:~} Beyond prediction, \method can leverage uncertainty in anomaly scoring by factoring in both prediction error as well as the uncertainty estimates, which allows it to flag abnormal behavior with greater accuracy. 
    
    

    \item \textbf{Effectiveness and Generality:}  We deploy \method at Novateur on industry-scale data and show that
     it improves AUROC performance by 3\%-15\% over various anomaly detection baselines. Although \method is designed with human mobility in mind, it is directly applicable to other industrial applications,  where user behavior modeling and anomaly detection are of utmost value.
     

\end{itemize}

%% file: tables/tab_related.tex
\begin{table}[!t]
	    \centering
	    \caption{Comparison between our model and related works. (Abbr; DES: discrete event sequence, UM: uncertainty modeling, AD:  anomaly detection, STD: spatiotemporal dependencies, MTF: Mixed-type, numeric\&categ,  features)}
            \vspace{-0.15in}
           \setlength\tabcolsep{1.5 pt}
    	\resizebox{0.9 \linewidth}{!}{
    		\begin{tabular}{ccccccc}
    			\toprule
    			Methods & DES & UM  & AD & STD & MTF\\
    			\midrule
    			  Human mobility modeling \cite{2021CTLE,2023LightPath,DeepMove,want2024trans,MobTCast}       & \ding{52} &   &   & \ding{52} & \ding{52}\\
                    Uncertainty learning in deep models \cite{kendall2017,Huang2018semantic,sensoy2018evidential,ye2024uncertaintyregularized,chanestimating}       &  & \ding{52} &  &  & \ding{52}\\
    			  Trajectory anomaly detection  \cite{IBAT2011, ATROM, song2018anomalous, onlineGMVSAE2020, li2024difftad} &  &  & \ding{52} & \ding{52} &   \\
                    Time series anomaly detection \cite{2024DualTF, Feng2024HUE, zamanzadeh2024deep, wang2024revisiting}  &  & \ding{52} & \ding{52} &  &  \\
                   
    			\midrule
    			\midrule
    			 \method (ours)  & \ding{52} & \textbf{\ding{52}} & \textbf{\ding{52}}  & \textbf{\ding{52}} & \textbf{\ding{52}} \\
    			\bottomrule
    		\end{tabular}
    	}
	\label{tab:method_compare}
\end{table}

%% file: 02prelim.tex

We introduce key concepts and formally define the anomaly detection problem in human mobility. Traditional trajectory-based anomaly detection methods process sequences of regularly sampled GPS readings in the form of $\langle x,y, t\rangle$. However, this formulation cannot capture the semantic meaning of human mobility patterns, e.g., the fact that a person typically spends an hour at lunch before returning to the office. To overcome this limitation, we transform raw GPS points into irregularly sampled stay-point events, which we formally define in the next paragraph. This transformation acts as a ``tokenization'' step, converting low-level GPS data into a structured, higher-level representation that enables the model to leverage rich semantic features more effectively.

\begin{figure}[!t]
 \vspace{-0.75em}
\includegraphics[width= 0.7\linewidth]{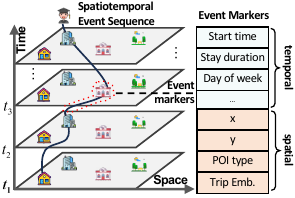}
   \vspace{-0.1in}
\caption{Spatiotemporal event sequence with various features (a.k.a. markers) per event.}
\label{fig:event_sequence}
    \vspace{-0.2in}
\end{figure}

\noindent \textbf{Stay-point Event}. As shown in Figure~\ref{fig:event_sequence}, a stay-point event (simply event) is derived from raw GPS data to represent an individual's daily activities. It is defined as a stationary point where GPS readings remain unchanged for at least five minutes. Let $e_i$ be an event with spatiotemporal features (or markers) $\bm x_i$:
\begin{equation}
    {\bm x_i} = (x_i^{\rm st}, x_i^{\rm sd}, x_i^{\rm x}, x_i^{\rm y}, x_i^{\rm poi}, x_i^{\rm dow}, {\bm x}_i^{\rm tripemb}) \;,
\end{equation} 
where $\F_n = \{x_i^{\rm st}, x_i^{\rm sd}, x_i^{\rm x}, x_i^{\rm y}\}$ is the numerical feature set: $x_i^{\rm st} \in \R$ and  $x_i^{\rm sd} \in \R$ are the start time and stay duration of the event; $x_i^{\rm x}, x_i^{\rm y}$ are the two-dimensional coordinates depicting 
the latitude and longitude of the event's location.  $\F_c = \{x_i^{\rm poi}, x_i^{\rm dow}\}$ denotes the categorical feature set; $x_i^{\rm poi} \in \mathcal{P}$ is the Point-of-Interest (POI) such as office, store, etc. with $|\mathcal{P}|=\#{\rm poi}$  unique POIs (see Table \ref{tab:stat_test_data}), and  $x^{\rm dow} \in \mathcal{D}$ depicts the Day-of-Week (DOW) with $|\mathcal{D}|=7$. 


\noindent \textbf{Trips.~} Besides stay events, humans are engaged with commutes (i.e. trips). There exists a trip between every two consecutive stay-points, depicted by a sequence of GPS coordinates over time. A trip trajectory conveys information about the shape and speed from one stay-point to a destination stay-point. We incorporate trip data by embedding these trajectories \cite{cao2025} and concatenating the embedding vector ${\bm x}_i^{\rm tripemb} \in \R^{D}$ to the markers of the destination event $e_i$, where $D$ is the embedding dimension. The goal is to learn the relationship between trips and stay events (e.g. certain routes taken to office) and the deviations thereof for better anomaly detection.


\noindent \textbf{Spatiotemporal Event Sequence.} Let ${\mathcal E}^u =  [e_1^u, e_2^u, \dots, e_{N_u}^{u}]$ be the sequence of all historical events recorded for an individual $u$ in the dataset, where $N_u$ is $u$'s total total number of historical events. Furthermore, let $e_i^{u,d}$ be the $i$-th event of individual $u$ on day $d$, and $N_{u,d}$ be the total number of events for $u$ on that day $d$. The event sequence with $w$-day time window can be given as
\begin{equation}
    {\mathcal E}^u_w =  [e_1^{u, d-w}, e_2^{u, d-w}, \dots, e_1^{u, d-w+1}, \ldots, e_1^{u, d}, \dots, e_{N_{u,d}}^{u, d}] \;.
\end{equation}



\noindent \textbf{Unsupervised Human Mobility Anomaly Detection.} Given an individual $u$'s event sequence over a $w$-day window ${\mathcal E}_u^{w}$ and a target event $e \in {\mathcal E}_u^{w}$, our goal is to design an anomaly score function that quantifies how anomalous $e$ is, i.e., how much it deviates from $u$'s typical behaviors, without relying on any labeled data.



%% file: 03method.tex
\par
\begin{figure}[!t]
    \centering
    \includegraphics[width=1\linewidth]{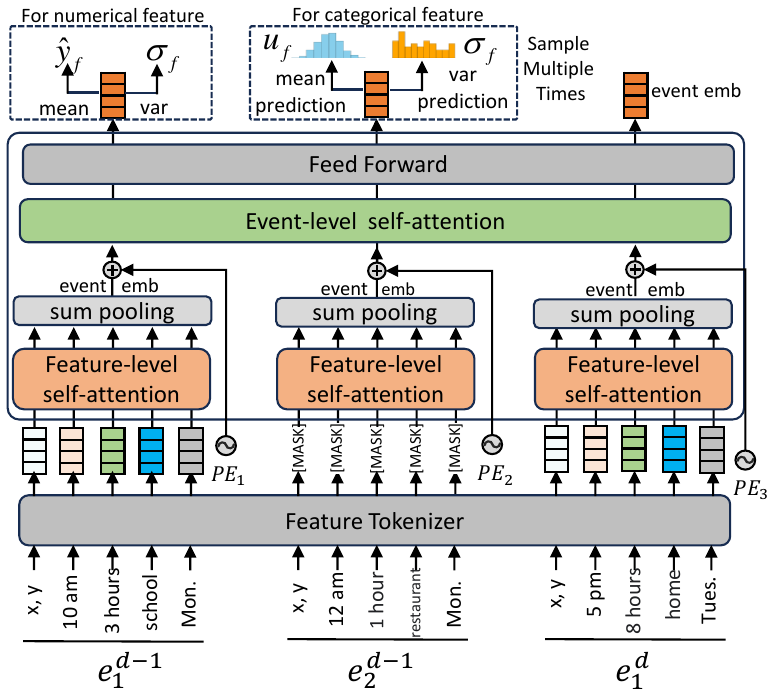}
    \caption{Proposed model architecture for uncertainty-aware human mobility behavior modeling. Raw GPS data is represented as a (ordered) sequence of stay-point events each with a (unordered) sequence of spatiotemporal markers. Dual-Transformer models each individual's sequence-of-sequences via both feature- and event-level attention, as well as uncertainty estimation that simultaneously enables robust training and informs anomaly scoring at inference.}
    \label{fig:model}
\end{figure}

 Figure~\ref{fig:model} illustrates the architecture of \method, 
which learns human mobility patterns in a self-supervised manner through pre-training, with the goal of accurately reconstructing masked events as well as quantifying the prediction uncertainty. In a nutshell, we first tokenize the input by projecting each individual feature of an event to a high-dimensional token. Then, a Dual-Transformer encoder encodes the input by both feature-level and event-level self-attention (\S\ref{ssec:encoder}). Unlike most prior work, we introduce an uncertainty-aware decoder to recover the masked event, to estimate the aleatoric and epistemic uncertainty of each feature (\S\ref{ssec:decoder}). Lastly, the estimated uncertainties are combined with the predicted error to form the uncertainty-incorporated anomaly scoring (\S\ref{ssec:ucanomaly}).

\vspace{-0.1in}
\subsection{Dual Transformer Encoder} 
\label{ssec:encoder}

\par Inspired by \citep{truong2023poet}, we design a Dual-Transformer encoder that treats the data as a \textit{sequence of sequences}, 
which takes the sequence of events each with a sequence of feature tokens as input, and models both feature-level and event-level interactions with two types of Transformer-based \citep{Transformer} components. \sloppy{Details of the Dual-Transformer block are included} in Appx~\ref{appendix:transformer}.

\textbf{Feature Tokenizer.} The feature tokenizer transforms all features ${\bm x} \in {\mathbb R}^{F} $ of an event into embeddings ${\bm e} \in {\mathbb R}^{F \times D}$, where $F$ is the number of input features and $D$ is the embedding dimension. Specifically, a numerical feature is projected by a linear transformation, while a categorical feature is projected by an embedding layer. We refer to Appx~\ref{appendix:feature_tokenizer} for more details. The trip information is encoded into embeddings (in Section~\ref{sec:prelim}) and treated as numerical input. Denoting $B$ as the batch size and $L$ the max length of ${\mathcal E}^u_w$ in a batch, the output of the feature tokenizer is $\E_0 \in \R^{B \times L \times F \times D}$.

\textbf{Feature-level Transformer.} By considering each feature as an input token to the Transformer, the Feature-level Transformer represents a feature in a way that explicitly incorporates other features' information. In the implementation, we convert the feature tokenizer's output $\E_0 \in \R^{B \times L \times F \times D}$ into $\E_{1} \in \R^{(B \times L) \times F \times D}$, which is then fed into $M_{1}$ Transformer blocks to get the updated feature embeddings $\E_{2} \in \R^{(B \times L) \times F \times D}$. Note that we do not include any positional embedding for this module, as there is no sequential or order relationship between the features of an event.

\textbf{Event-level Transformer.} 
Based on $\E_{2}$, we calculate an event's embedding by summing all its feature embeddings, which results in the input of the Event-level Transformer $\E_{3} \in \R^{B \times L \times D}$. The Event-level Transformer then captures the dependencies (e.g., sequential patterns) between different events by considering each event as a token. Moreover, considering that the sequential information is important to reflect human mobility behavior, we design two types of positional encoding for each event: ($i$) Sequence positional encoding;  used to describe the order of an event in the input sequence,  and ($ii$) Within-day positional encoding; used to describe the order of the event in its specific day, explicitly demarcating day boundaries. After $M_{2}$ updates of the Event-level Transformer blocks, 
we get the input of the decoder $\bar {\E}  \in \R^{B \times L \times D}$.

\vspace{-0.75em}
\subsection{Uncertainty-aware Mobility Learning} 
\label{ssec:decoder}

\par \textbf{Motivation.~}  Human mobility sequences are inherently associated with uncertainty. For instance, certain individuals (e.g. shift workers) may exhibit highly predictable patterns (i.e. low uncertainty), while others (e.g. retirees) can be much more unpredictable in their activities over time. Such uncertainty, known as aleatoric uncertainty (AU) \citep{kendall2017}, arises naturally from the data itself and cannot be reduced by simply adding more training data. In addition to AU, human event sequence modeling also involves epistemic uncertainty (EU), which refers to the uncertainty in the model itself due to limited knowledge. EU captures the model’s confidence in its predictions and can be reduced by acquiring more data or improving the model specification or architecture. Estimating the underlying uncertainty helps achieve more accurate mobility modeling, which further informs anomaly scoring at inference.

As various features can have different predictability even for the same individual or event, we propose to estimate the uncertainty measurement at the feature level.
 Besides, unlike most previous works which study AU and EU only for numerical or categorical input \cite{amini2020deep,sensoy2018evidential,ye2024uncertaintyregularized,chanestimating}, here we incorporate the uncertainty of both types in a single and unified architecture, to offer substantial real-world relevance and industrial applicability.


\vspace{-0.75em}
\subsubsection{Epistemic Uncertainty (EU) Modeling}
\label{ssec:epis}
We model EU by placing a prior distribution over the model's parameters \citep{kendall2018phd},  then try to capture how much these parameters vary given observed data. 

To ease the presentation, we use $s$ to denote the input instance ${\mathcal E}_w^u$, and $\theta$ to denote all parameters of the model $\mathcal{M}$. We start by assuming $\theta$ follows a prior distribution, which reflects our initial belief about the possible values of these parameters. The goal is to update this belief based on the available data $s$ by calculating the posterior distribution of $\theta$, that is $p(\theta | s)$. This posterior captures how the parameters might vary, thereby reflecting the model's uncertainty about its predictions. Formally, Bayes' theorem \citep{bayes1763} gives us the posterior as $p(\theta | s) = \frac{p(s | \theta) p(\theta)}{p(s)}$.


However, computing the exact posterior distribution in deep learning models is intractable due to the high dimensionality of the parameter space and the complexity of the model. As a practical alternative, we apply Monte Carlo (MC) Dropout \citep{gal2016dropout} to approximate the posterior, with dropout ratio 0.05.  MC Dropout samples different subsets of the model parameters $\theta$ during both training and inference, thereby creating an ensemble of models. The variance in the predictions across these passes provides an estimate of the epistemic uncertainty. 



\textbf{\textit{Epistemic uncertainty for numerical features:~}} Decoding the numerical feature $f \in \F_n$ is modeled as a regression task. In this case,  the variance of predictions across $T$ (equals 50 in our model) steps of stochastic forward passes quantifies the epistemic uncertainty $\alpha_f^{num}$ for $f$, formulated as 
\vspace{-0.1in}
\begin{equation}
    \alpha_f^{num}:= \frac{1}{T} \sum_{t=1}^{T} \left(\M_{\theta_t}(s) - \bar{y}\right)^2 ,  {\rm where~} \bar{y} = \frac{1}{T} \sum_{t=1}^{T} \M_{\theta_t}(s) \;.   
    \label{eq:num_EU}
\end{equation}
\vspace{-0.1in}

\textbf{\textit{Epistemic uncertainty for categorical features:~}} Decoding categorical feature $f \in \F_c$ is modeled as a classification task. Given the average probability distribution over $T$ forward processes: ${p(y|s)} = \frac{1}{T} \sum_{t=1}^{T} {\rm softmax}(\M_{\theta_t}(s))$, where softmax is applied to convert logits into probability each time, EU is measured by the entropy of the average predicted probability distribution:
 \vspace{-0.25em}
\begin{equation}
        \alpha_f^{cls}: H(p) = - \sum_{c} {p}_c \log(p_c), {\rm~where~} {p}_c=p(y=c|s) \;.  
    \label{eq:cls_EU}
\end{equation}






\vspace{-1em}
\subsubsection{Aleatoric Uncertainty (AU) Modeling} 
\label{ssec:alea}
We model aleatoric uncertainty by placing a distribution over the model's output (decoded based on embeddings $\bar{\E}$). Specifically, we model the output as a Gaussian distribution with random noise, i.e., $ p(y | \M_{\theta}(s)) = \mathcal{N}(\M_{\theta}(s), \sigma^2(s))$, where AU aims to estimate the variance as heteroscedastic (i.e. input-dependent) data noise.


\par \textbf{\textit{Aleatoric uncertainty for numerical features:~}} Let $\sigma_f$ denote the noise level of a feature $f \in \F_n$. As  we assume $\sigma_f$ can be data-dependent, \method  estimates it at the output as a function of the input data $s$, which serves as a learned loss attenuation:
\vspace{-0.35em}
\begin{equation}
\mathcal{L}_f^{num}=\frac{1}{2 \sigma_f^2(s)}\left\|y_f-\widehat{y}_f\right\|^2+\frac{1}{2} \log \sigma_f^2(s) \;,
\label{eq:reg_loss}
\end{equation}

\vspace{-0.5em}
\par \noindent where $y_f$ and $\widehat{y}_f$ are the true and predicted value.  Notice that a higher $\sigma_f$ lowers the penalty of the error term when the input data is associated with high aleatoric uncertainty, thus making the loss more robust to noisy data. The second term is for regularization that prevents the model from learning a high uncertainty score for all instances, trivially driving the first term to zero. In practice, we  use  
$ \mathcal{L}_f^{num}=\frac{1}{2} \exp \left(-r_f\right)\left\|y_f-\hat{y}_f\right\|^2+\frac{1}{2} r_f$,  
where $r_f=\log \sigma_f^2(s)$, as predicting the log-variance 
ensures the estimated variance has a positive value that makes the training more numerically stable. 
Overall, at inference the aleatoric uncertainty for numerical feature $f$ is calculated as $\beta_f^{num}:= \frac{1}{T}\sum_{t=1}^T \sigma_{f,t}^2(s)$.

    \textbf{\textit{Aleatoric uncertainty for categorical features:~}} Let $C_f$ denote the number of unique categories (i.e. classes) of feature $f \in \F_c$. For classification, \method assumes that the prediction logits at sample time $t$ for each category follows a  Gaussian distribution:
\vspace{-0.25em}
\begin{equation}
    \widehat{\bm{m}}_{f, t}=\bm{u}_{f}+\bm{\sigma}_{f} \cdot {\bm \epsilon_t}, \quad {\bm \epsilon}_t \sim \mathcal{N}(0, I) \;,
\end{equation}
\vspace{-1.5em}

\par \noindent where the predicted mean logits $\bm{u}_{f} \in \R^{C_f} $ and variance ${\bm \sigma}_{f} \in \R^{C_f}$ are model outputs as a function of the input with sampled parameters at $t$, and ${\bm \epsilon}_t \in \R^{C_f}$ represents a random vector drawn from a unit normal distribution $\mathcal{N}(0,I)$. Here ${\bm \sigma}_{f}$ accounts for the inherent uncertainty in the feature when conducting the prediction. At inference, the aleatoric uncertainty of categorical feature $f$ is calculated as the variance of the \textit{true} class $c$, as $\beta_f^{cls}:= \frac{1}{T}\sum_{t=1}^T{\sigma}_{f, t,c}^{2}$.


To learn such uncertainty, the loss function $\mathcal{L}_{f}^{cls}$ calculates the cross entropy over the average output probability (calculated by softmax over the logits $\widehat{{m}}_{f, t}$) across $T$ steps:


\vspace{-0.175in}
\begin{equation}
    \mathcal{L}_{f}^{cls}=\log \frac{1}{T} \sum_t \exp \left(\widehat{{m}}_{f, t, c}-\log \sum_{c^{\prime}} \exp \widehat{{m}}_{f, t, c^{\prime}}\right).
    \label{eq:cls_loss}
\end{equation}
When the model assigns a high logit value  $m_c$  to the true class $c$, and the noise $\sigma_c$  is low, the loss approaches zero — which is the desired outcome.  On the other hand, if the model cannot learn the true class, $\sigma_c$ becomes high to attenuate the loss. As such, 
the estimated uncertainty provides robust training against noisy input, similar to the numerical case \citep{kendall2017}.  Overall, the total loss is the sum of the regression and classification losses across all features:
\begin{equation}
    {\mathcal{L}} = {\sum}_{f \in \F_n}{\mathcal L}_{f}^{num} \;+\; \lambda {\sum}_{f \in \F_c}{\mathcal L}_{f}^{cls} \;,
\end{equation}
where $\lambda$ is a hyper-parameter to balance the scale of regression and classification losses. In our model, we set $\lambda=1$ for simplicity.


\subsection{Uncertainty-incorporated Anomaly Scoring}
\label{ssec:ucanomaly}

Until now, we have learned inherent uncertainties in human mobility, which raises critical questions: 
How can uncertainty be effectively incorporated into anomaly scoring? While previous methods often rely solely on prediction error (PE), how can we leverage the combined insights from PE, AU, and EU? Moreover, what do AU and EU capture about the underlying behavior, and how can these insights guide the design of uncertainty-aware anomaly detection? In this section, we address these questions by examining the roles of AU and EU and their integration into anomaly scoring.

\begin{figure}[htbp]
  \includegraphics[width= \linewidth]{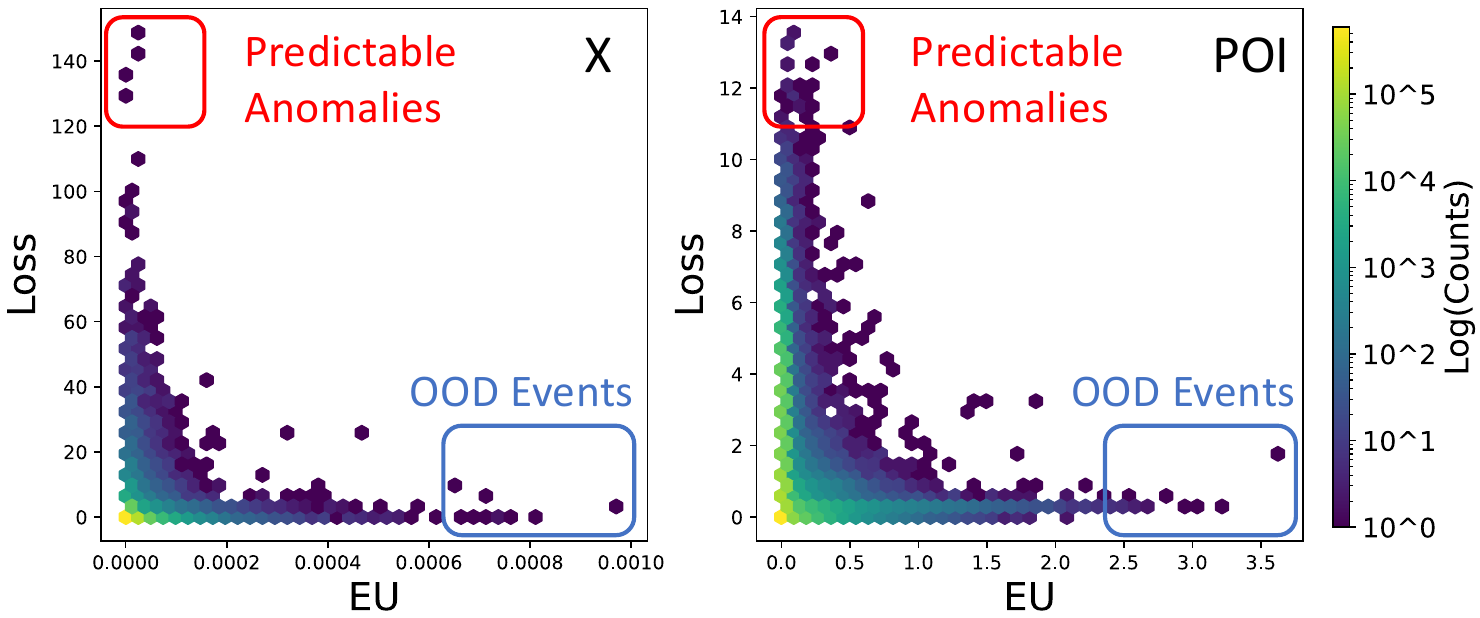}
   \vspace{-0.25in}
  \caption{Loss (AU-attenuated PE) vs. EU for features x (left) and POI (right). Color depicts log. of event counts w/ most points near origin. Two regions of interest are: $i$) predictable anomalies w/ low-EU\&high loss; $ii$) OOD events w/ high-EU. 
  }
  \label{fig:loss_eu}
\end{figure}


We start by plotting the loss (i.e., AU-attenuated PE) vs. EU for each event in Figure~\ref{fig:loss_eu}, for two example event features; $x$ (numerical) and POI (categorical).\footnote{The plots for other features are similar, and can be found in Appx. Figure~\ref{fig:loss_eu_all_features}.}
The loss for numerical features is defined as  $\mathcal{L}^{\prime}_f = \frac{1}{2 \beta_f^{num}} \left\| y_f - \widehat{y}_f \right\|^2 $, where \(y_f\) is the observed value, \(\widehat{y}_f\) is the predicted value, and $\beta_f^{num}$ is the estimated AU of the feature. For categorical features, the loss is as given in Eq.~\eqref{eq:cls_loss}. 

We can see that most samples cluster near the origin, with low loss and low EU, while two distinct regions of interest emerge: one with high loss and low EU, and vice versa. The first region corresponds to what we call ``predictable anomalies'', where the model is highly confident (low EU) but incurs large prediction error (high loss). These anomalies are indicative of events that deviate significantly from expected patterns within the model's knowledge. The second region is characterized by low loss but high EU associated with high-variance predictions.
It is not directly clear, however, how EU should factor into our anomaly scoring, if at all.




In investigating what the high-EU events reflect, 
we conjecture that high variance might stem from training data scarcity, that is, high EU predictions are those out-of-distribution (OOD) instances unseen in the training data. To that end, 
we conduct a detailed analysis (see Appx.~\ref{appendix:uncertainty}) on the relation between EU and kNN distance (i.e. average of the Euclidean distances between the test event and its $k$ nearest training events in the embedding space of \method) as a measure of OOD-ness. Our findings reveal that: $i$) Events with very high EU often exhibit large nearest-neighbor distances in the embedding space of our \method, suggesting they are dissimilar to the training data (see Appx. Figure~\ref{fig:eu_knn_hist}). However, the overall correlation between EU and kNN distance is weak (see Appx.~Figure~\ref{fig:eu_knn_scatter}). $ii$) Manually injected OOD events are not effectively detected by EU scores alone. Based on these insights, we exclude EU from the anomaly scoring as it reflects unreliable predictions. Instead, we directly incorporate kNN distance as a more effective metric to identify OOD anomalies (see Appx.~Table~\ref{tab:ablation_injection_1}). 
Overall, the anomaly score (AS) of an event $e$ is defined as

\vspace{-1.5em}
\begin{equation} 
\text{AS}(e) = \max \bigg\{  \max_{f \in \mathcal{F}_n \cup \mathcal{F}_c} \big\{\mathcal{L}_f^{\prime} \big\}, \;\text{kNN}(e) \bigg\} \;, 
\end{equation}

\vspace{-0.5em}
\par \noindent which is the maximum between the highest loss among the event features and the $k$-nearest neighbor distance of the event in the embedding space.\footnote{We apply percentile transform \cite{percentile} to each term before aggregation to make them comparable in scale.} 
As such, the anomaly score is driven by our earlier analyses, capturing both predictable anomalies (via prediction loss, i.e. the first term) as well as novel behavioral events (via kNN distance, i.e. the second term). As shown in our ablation studies (Figure~\ref{fig:score_ablation}), this scoring function outperforms other approaches such as using prediction loss alone or simply summing prediction error with AU and EU,  without carefully accounting for their distinct roles. The agent-level anomaly score is defined as the maximum event score across all events associated with a given agent.

\noindent \textbf{\textit{Complexity Analysis}}: Given a sequence of $L$ events, each with $F$ features. The time complexity in the feature-level self-attention attention layer is $O(F^2 \cdot D)$, and the complexity in the event-level self-attention layer is $O(L^2 \cdot D)$. Let $T$ denote the sample times at the inference process, the total complexity of the proposed model is $O(T(F^2 \cdot D + L^2 \cdot D))$. The model takes approximately 3 minutes to train one epoch (411k samples of batch size 128) and performs inference at an average speed of 0.04 seconds per sample for $T=50$ and $w=3$ on one GPU (Nvidia RTX A6000). Thus, it leads to 25 QPS (query per second) for batch size 1 for the deployed system, which can meet the speed needs of numerous real-world applications.

\noindent \textbf{\textit{Broader Impact for Industry}}. While \method is primally driven by human mobility data, it can be easily generalized for user behavior modeling in various application domains such as finance and e-commerce. We refer to Appx.~\ref{appendix:broader_impact} for further discussion.

%% file: 04experiments.tex

We \textbf{deploy} the solution at Novateur\footnote{https://novateur.ai/} and conduct extensive experiments to answer the following questions: \textbf{Q1: Prediction} What is the \method's performance in predicting masked-event features?  \textbf{Q2: Anomaly detection} Does uncertainty-aware \method outperform previous approaches in anomaly detection? \textbf{Q3: Uncertainty quantification} Does \method~have the ability to accurately capture  aleatoric and epistemic uncertainty? \textbf{Q4: Ablation study} How does various modules contribute to \method's performance? \textbf{Q5: Case studies} How does \method~help us understand and interpret the detected behavior anomalies?

 \textbf{Datasets.} We conduct experiments on 3 industry-scale datasets --- \trial, \trialthree, and \NUMOSIM --- each containing tens of thousands of agents and millions of events simulated in real-world cities. Each dataset is divided into two disjoint activity periods: The training period consists exclusively of normal activities, allowing the model to learn typical behavioral patterns. The testing period includes a small fraction of anomalous activities, enabling model evaluation to detect deviations from learned patterns. 
 
 \NUMOSIM \citep{numosim} is a public dataset for human mobility anomaly detection with randomly inserted anomalies. In contrast, \trial~ and \trialthree are private datasets where anomalies are expert-injected by independent ``red teams'' at Novateur. This ensures that $i$) the anomalies are realistic, and $ii$) \textit{our modeling remains agnostic to the way anomalies were introduced}. Each dataset includes two types of anomalies: (1) Non-recurring anomalies --- one-time deviations from an agent’s usual behavior, and (2) Recurring anomalies --- systematic behavioral shifts over time, simulating longer-term deviations from established patterns. In summary,
  \vspace{-1.5em}
\begin{itemize}[leftmargin=*]
    \item \NUMOSIM is simulated in Los Angeles with two months of mobility data for 20K agents, of which 381 (1.9\%) are anomalous, and 3.43 million events, with 3,468 (0.1\%) marked as anomalies.
    \item \trial is also simulated in Los Angeles with two-month mobility data for 20K agents, of which 162 (0.8\%) are anomalous, and 3.11 million events, with 892 (0.03\%) labeled as anomalous.
    \item \trialthree is simulated in Jordan, containing 20K agents, of which 370 (1.9\%) are anomalous. Unlike the other datasets, it includes only agent-level anomaly labels.
\end{itemize}
Detailed dataset statistics are provided in Appx. Table~\ref{tab:stat_test_data}, and Appx. \ref{appendix:data_simulation} discusses the motivation and methodology behind the anomaly simulation and injection process. Each dataset is split into train/validation/test by date using a 3/1/4 week ratio. During training, we randomly mask events in each sequence based on a tunable mask ratio, validated as a hyperparameter. For validation and testing, we mask one event at a time, as anomaly detection operates by scoring each incoming event over time.



\subsection{Q1: Prediction Performance}

\par \textbf{Baselines.} We include several deep models for human mobility modeling: MLP \citep{MLP}, LSTM \citep{LSTM}, Transformer \citep{Transformer}, LightPath \cite{2023LightPath}, and Dual-Tr which is the uncertainty-free variant of \method that only keeps the Dual Transformer architecture. More details are in Appx~\ref{appendix:masked_prediction_baselines}. In addition, since \method~can estimate the uncertainty of its prediction, we also report its prediction-with-rejection performance where the top-5\% most uncertain test samples are excluded from the evaluation, denoted by \methodwrej.

\par \textbf{Setup.} We use MAE and MAPE to evaluate the performance of numerical features and use accuracy (ACC) for categorical features.  The models are trained on a NVIDIA RTX A6000 GPU.  We search the hyperparameters in a given model space for all deep models, and each model's best performance is reported. Detailed settings and model configurations can be found in Appx.~\ref{appendix:masked_prediction_settings}.




\input{tables/tab_masked_prediction_in_appendix}

\par \textbf{Results.} 
In Table~\ref{tab:masked_prediction_big}, \method~ delivers the best performance for all features compared with the baselines. The results in \trialthree are similar and can be found in  Appx.~Table~\ref{tab:masked_prediction_trial3}.
Compared to Dual-Tr in \trial, \method reduces MAE  from 20.62 to 8.78 minutes for ST, and from 76.53  to 36.68 for SD, underscoring the significant impact of modeling both AU and EU in achieving more accurate predictions. Notably, \methodwrej further improves over \method, which demonstrates its ability to quantify uncertainty and that it can be quite accurate on its certain predictions.
Full results on prediction with 
varying certainty thresholds are shown  in Figure~\ref{fig:uc_acc_relation}.



\subsection{Q2: Anomaly Detection Performance}
\par \textbf{Baselines.}  For comprehensive evaluation, we employ baselines from three categories: (1) Human mobility learning:  Transformer \cite{Transformer} and  CTLE \cite{2021CTLE}; (2) Trajectory anomaly detection: IBAT \cite{IBAT2011}, GMVSAE \cite{onlineGMVSAE2020}, ATROM \cite{ATROM}; and (3) Multi-variate time series anomaly detection: SensitiveHUE \cite{Feng2024HUE}.  Details are in Appx \ref{appendix:anomaly_detection_baselines}. 

\par \textbf{Setup.} We evaluate both event-level (if applicable) and agent-level anomaly detection by two common metrics, AUROC (Area Under ROC) curve and AUPR (Area Under Precision-Recall) curve. More detailed settings (including parameter and anomaly scoring of baselines) can be found in Appx.~\ref{appendix:anomaly_detection_setting}.
Note that \trialthree does not have event-level anomaly labels, and baselines IBAT, GMVSAE and ATROM do not have the ability to flag event-level anomalies.


\par \textbf{Results.}
As shown in  Table~\ref{tab:anomaly_detection}, \method significantly outperforms the human mobility modeling methods like CTLE and the uncertainty-free Transformer, demonstrating the effectiveness of incorporating both aleatoric and epistemic uncertainty. 
Trajectory anomaly detection methods (IBAT, GMVSAE, ATROM) are limited to agent-level detection and perform poorly as they rely solely on simple GPS coordinates, neglecting the complex spatio-temporal behaviors and semantic meaning inherent in human mobility data.
\sloppy{SensitiveHUE is primarily designed to detect anomalies in multivariate time series where it performs better than previous methods by accounting for uncertainty. However, it is limited to only estimating the temporal dependencies, failing to capture the spatial and temporal interactions in human mobility.
Moreover, SensitiveHUE's inability to process categorical features further limits its performance compared to \method. 

}

\input{tables/tab_AD_new}



\vspace{-0.1in}
\subsection{Q3: Analysis of Uncertainty Estimation}

Next, we analyze the quality of the estimated uncertainties, by examining their relationship to the prediction performance.




\begin{figure}[bp]
    \centering
    \vspace{-1.5em}
    \includegraphics[width= 1 \linewidth]{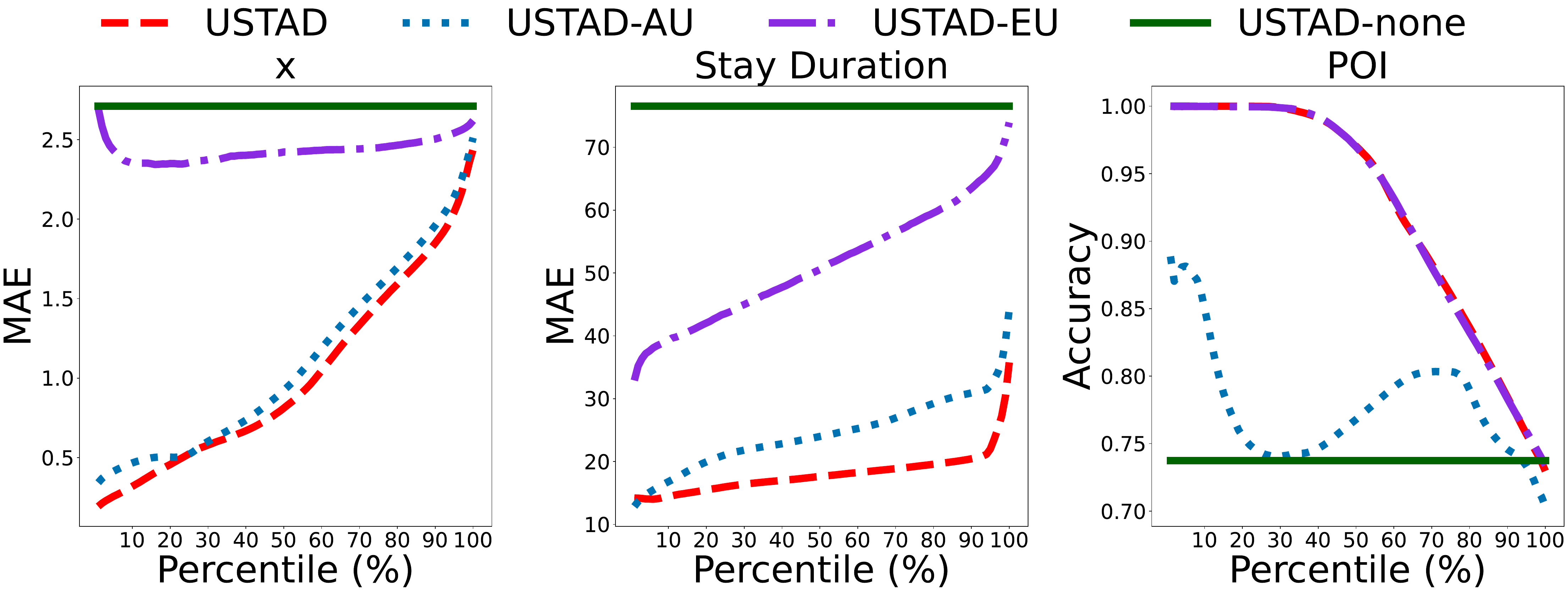}
    \vspace{-2em}
    \caption{Relationship btwn. MAE (accuracy) \& uncertainty follow an increasing (decreasing) trend. The uncertainty estimates are well aligned with prediction performance.}
    \vspace{-1em}
\label{fig:uc_acc_relation}
\end{figure}

\textbf{Relationship between uncertainty and MAE \& Accuracy.} We first sort the test samples by their total uncertainty (AU+EU, i.e., $\sum_{f\in \F}\alpha_f + \beta_f$), then gradually remove samples with uncertainties above certain percentile thresholds, and record \method's prediction performance on the remaining samples. As in Figure~\ref{fig:uc_acc_relation}, \method's performance improves as test samples with higher uncertainty are excluded (i.e., as the percentile threshold decreases). This indicates that the uncertainty estimates align well with prediction performance. Specifically, for numerical features, MAE curves show a monotonically increasing trend, while for categorical features, the accuracy curve exhibits a decreasing trend.
Notably, for stay duration prediction (middle), MAE increases sharply from around 20 to 35 minutes due to the top 5\% most uncertain samples. For POI prediction (right), the most certain 50\% of samples achieve nearly 100\%  accuracy, which gradually declines as more samples with higher uncertainties are included.





\begin{figure}[htbp]  \includegraphics[width= 0.6\linewidth]{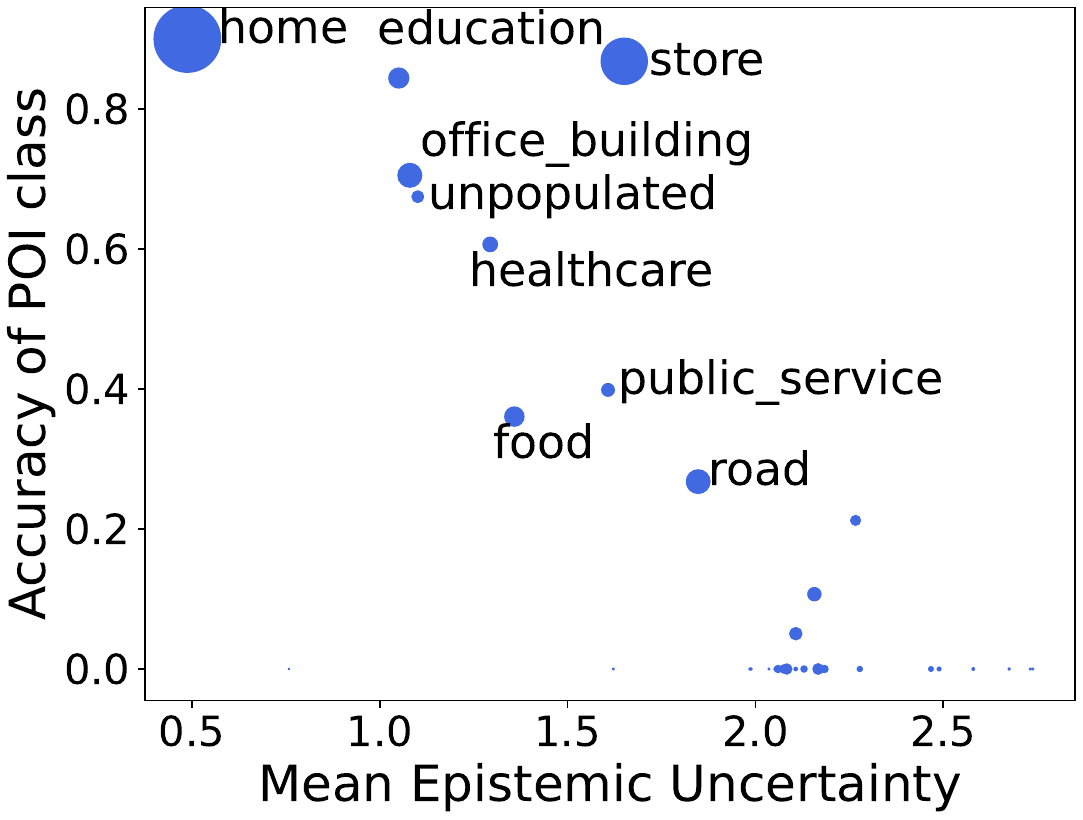}
   \vspace{-1em}
  \caption{Relationship between EU and accuracy of each POI type (only a few are annotated for better illustration). \method accurately captures uncertainty: POI prediction performance tends to increase when EU is lower. The size of each point/POI is proportional to its frequency in the dataset. }
  \label{fig:EU_POI_relation}
  \vspace{-1em}
\end{figure}

\textbf{Relationship between EU and accuracy of each POI class}. From Figure~\ref{fig:EU_POI_relation}, we have the following observations: ($i$) \textit{Inverse correlation:} As \method becomes more certain (lower EU), its prediction accuracy tends to increase. ($ii$) \textit{Impact of POI frequency:} Higher frequency POI classes (e.g., home and office buildings) are associated with lower EU and higher accuracy. This is expected since these classes have more training samples, leading to better model performance. ($iii$) \textit{Exceptions:} A lower frequency does not always result in a high EU. For instance, POI:home and POI:education both have relatively low EU (and similarly high accuracies), while the latter has a much lower
frequency. The low EU for POI:education is possibly because people's visits to such POI are highly regular. 



\vspace{-0.5em}

We further analyze how AU and EU estimates change as we vary the amount of training data. Detailed results in Appx. Table \ref{appendix:tab:uc_train_data_relation} shows that, as expected, EU decreases while AU remains nearly constant as the size of the training dataset increases.

\subsection{Q4: Ablation Study}



\textbf{Ablation on Uncertainty Estimation}. 
To evaluate the role of AU and EU estimation in \method, we compare three variants: ($i$) \method-AU, which only implements the AU estimation while keeping the model backbone unchanged, ($ii$) \method-EU,  which only models EU; and ($iii$) \method-none, which only has the bare Dual Transformer and does not model uncertainty. For comparison, Figure~\ref{fig:uc_acc_relation} plots the relationship between uncertainty and performance. Uncertainty modeling significantly enhances the prediction accuracy of numerical features. Interestingly, \method-AU ({in blue}) appears to be more effective in improving performance for numerical features (x and stay duration), while it is the EU-only variant (in purple) for categorical features (POI), demonstrating the complementary benefit of estimating these different uncertainty types.






\begin{figure}[!t]

  \includegraphics[width= 0.95\linewidth]{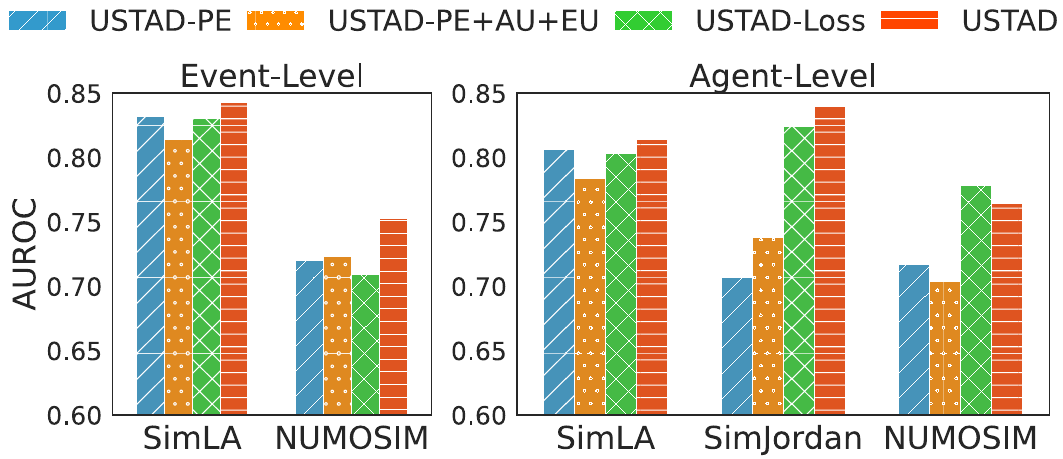}
   \vspace{-1em}
  \caption{Comparison of different anomaly scores.  \method  (red) outperforms prediction error (PE) alone (blue), loss alone (green), and simple sum of PE+uncertainties (orange).} 
  \label{fig:score_ablation}
  \vspace{-2em}
\end{figure}

\begin{figure*}[!t]
\vspace{-0.1in}
    \centering
    \includegraphics[width=1 \linewidth]{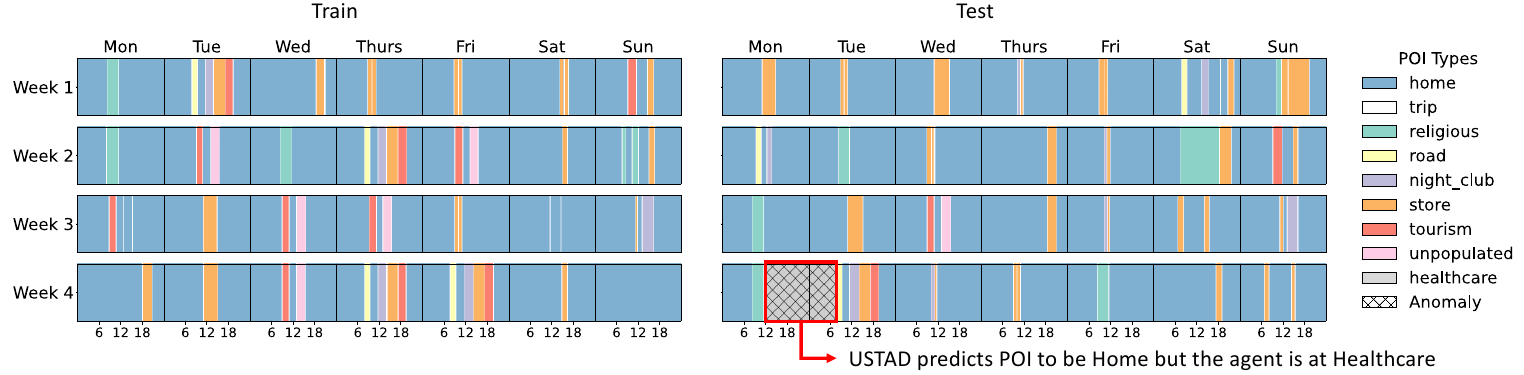}
    \caption{(best in color) Case study: An \method-detected ``predictable anomaly'' (i.e. high loss \& low EU)
    based on POI. An agent's behavior during train (left) and test (right) periods. From noon to early morning, s/he usually stays at Home (blue), with an unexpected deviation to a Healthcare facility (gray) in Week 4. Here, \method makes a confident (i.e. low EU) POI prediction as Home, but with the observed POI being Healthcare (thus incurring high loss).}
    \label{fig:high_poi_loss_main}
\end{figure*}

\textbf{Ablation on Different Anomaly Scores}. Figure~\ref{fig:score_ablation} compares the performance of \method's anomaly detection using four different scores. Compared to PE (in blue), loss (in green), which incorporates AU,  generally improves performance, especially for agent-level detection. This suggests that AU helps distinguish between anomalous versus high-noise samples, where the latter may reflect stochastic behaviors rather than true anomalies. The hybrid score by \method (in red), utilizing loss and kNN distance,  generally outperforms loss alone, especially for \trial and \trialthree, indicating its effectiveness in capturing OOD events. Additionally, it consistently outperforms PE+AU+EU (in orange) which suggests that explicitly capturing predictable anomalies (by loss) and OOD events (by kNN distance) is more effective than simply combining prediction error with AU and EU, without considering their distinct roles. The results on AUPR are similar and can be found in~\ref{appendix:anomaly_scores}.
\subsection{Q5: Case Studies}

To gain an intuitive understanding of the anomalies detected by our model, we examine two representative cases: one with high loss and low EU (what we called a predictable anomaly), and one with high kNN distance (an OOD event). Figure~\ref{fig:high_poi_loss_main} presents the case of a predictable anomaly, where we visualize the agent’s event sequences during the training and testing periods to illustrate the expected behavioral patterns and the input context of the target anomaly event. 
The agent demonstrates a regular behavioral pattern during the training period, typically remaining at home from noon to early morning. Given this consistency, \method~confidently (i.e. with low EU) predicts the expected behavior. However, the agent deviates from this routine by visiting a healthcare facility, a pattern not observed in the training data. This unexpected deviation incurs a high prediction loss, thereby triggering anomaly detection.
For other cases including OOD events, we refer to Appx.~\ref{appendix:more_cases}. In summary, \method effectively utilizes uncertainty to detect anomalies. Further, the estimated uncertainties contribute to the interpretability by reflecting underlying data characteristics.

%% file: tables/tab_masked_prediction_in_appendix.tex
\begin{table*}[htbp]
\centering
\small
\caption{Masked prediction results. \method outperforms the baselines thanks to robust training w/ uncertainty modeling,  while allowing conformal prediction: \methodwrej capitalizes on its uncertainty estimates for prediction-w/-rejection (top 5\% most uncertain). Best performance in \textbf{bold}, 2nd best \underline{underlined}. Abbr: x:lat. , y:long. (km), ST: start time (min), SD: stay dur. (min). }
\vspace{-0.1in}
\renewcommand\arraystretch{1.1}
\setlength\tabcolsep{2 pt}
\resizebox{1.0 \textwidth}{!}{
\begin{tabular}{c|cccc|cccc|c||cccc|cccc|c}
\toprule 
\multirow{3}{*}{Method}
& \multicolumn{9}{|c||}{\trial}& \multicolumn{9}{|c}{\NUMOSIM} \\
\cline{2-19}
& \multicolumn{4}{|c|}{MAE $\;\;\downarrow$ }& \multicolumn{4}{|c|}{ MAPE (\%) $\;\;\downarrow$}& \multicolumn{1}{|c||}{ACC(\%) $\;\uparrow$} & \multicolumn{4}{|c|}{MAE $\;\;\downarrow$}& \multicolumn{4}{|c|}{MAPE (\%) $\;\;\downarrow$}& \multicolumn{1}{|c|}{ACC(\%) $\;\uparrow$}\\
\cline{2-19}
 & x & y & ST & SD &  x & y & ST & SD & POI & x & y & ST & SD & x & y & ST & SD & POI \\
\midrule 

MLP~\citep{MLP}  & 9.73 & 10.16 & 208.07 & 420.92 & 188.13 & 500.59 & 43.28 & 569.01 & 45.61 & 15.97 & 15.74 & 207.64 & 396.58 &  416.55& 361.58 & 339.15 & 572.69 & 56.08  \\ \hline

LSTM~\citep{LSTM} & 4.73 & 4.63 & 19.85 & 322.62 & 204.73 & 195.76 & 2.48 & 420.76 & 63.32 & 6.92 & 6.40 & 34.41 & 343.82 & 265.52 & 139.71 & 27.03 & 547.02 & 68.29  \\ \hline 

Transformer~\citep{Transformer} & 2.67 & 2.65 & 26.85 & 77.12 & 132.33 & 115.35 & 2.48 & 66.82 & 73.49 & 4.08 & 3.45 & 21.57 & 71.33 & 143.81 & 72.15 & \textbf{1.94} & 68.51 & 75.90  \\ \hline




LightPath~\cite{2023LightPath} &  2.66  &  2.91  &  34.90  &  143.73  &  133.04  &  125.41  &  2.22  &  138.91  &  -  & 4.65  &  4.63  &  54.47  &  138.65  &  186.42  &  108.76  &  30.1  &  200.93  &  - \\ \hline

Dual-Tr & 2.71 & 2.63 &  20.62 & 76.53 & 131.72 & 117.29  & 2.70 & 51.95 & 73.74 & \underline{3.63} & 3.48 & 21.69 & 80.08 & 134.32 & 71.34 & \underline{2.21} & 60.17 & \textbf{76.12}\\

\midrule \hline

\method & \underline{2.21} & \underline{1.94}  & \underline{8.78} & \underline{36.68} & \underline{105.65} & \underline{91.09} & \underline{1.98} & \underline{22.44} & \underline{74.14} & 3.65 & \underline{3.36} & \underline{10.40} & \underline{45.36} & \underline{120.81} & \underline{71.14} & 6.11 & \underline{42.87} & 74.82 \\

\methodwrej & \textbf{1.57} & \textbf{1.42} & \textbf{6.44} & \textbf{20.51} & \textbf{84.47} & \textbf{81.22} & \textbf{1.36} & \textbf{20.30} & \textbf{76.05} & \textbf{3.31} & \textbf{2.97} & \textbf{8.02} & \textbf{27.77} & \textbf{116.53} & \textbf{52.90} & 5.05 & \textbf{35.23} & \underline{76.01} \\

\bottomrule

\end{tabular}
}
\label{tab:masked_prediction_big}
\vspace{-1em}
\end{table*}

%% file: tables/tab_AD_new.tex
\begin{table*}[!t]
    \caption{Anomaly detection results (mean $\pm$ standard dev. over five seeds). \method significantly outperforms all baselines at both event-level (left) and agent-level (right) anomaly detection by 3.4\% up to 148.3\% w.r.t. both AUROC and AUPR performance. Improvement is calculated by: $|a - b| / b$, where $a$ is the metric of \method, and $b$ is the metric of the most competitive baseline. Dash (-) denotes the cases for the baselines that cannot provide event-level but only agent-level detection.} 
    \vspace{-0.1in}
    \centering
    \renewcommand\arraystretch{1.1}
    \setlength\tabcolsep{2 pt}
    \resizebox{1 \linewidth}{!}{
    \begin{tabular}{c|cc|cc||cc|cc|cc}
    \toprule
        \multirow{2}{*}{Method} & \multicolumn{4}{c||}{Event-level} & \multicolumn{6}{c}{Agent-level}  \\
        \cline{2-11}
         & \multicolumn{2}{c|}{\trial}& \multicolumn{2}{c||}{\NUMOSIM} & \multicolumn{2}{c|}{\trial} & \multicolumn{2}{c|}{\trialthree} & \multicolumn{2}{c}{\NUMOSIM}  \\
        \cline{2-11}
         & AUROC & AUPR$(10^{-1})$ & AUROC & AUPR$(10^{-1})$ & AUROC & AUPR$(10^{-1})$ & AUROC & AUPR$(10^{-1})$ & AUROC & AUPR$(10^{-1})$ \\
         \midrule
        CTLE \cite{2021CTLE}  &  $0.716\pm0.001$  & $0.061\pm0.001$  & $0.616\pm0.001$ &  $0.056\pm0.001$ & $0.561\pm0.005$ & $0.268\pm0.024$ & $0.656\pm0.001$ & $0.345\pm0.010$ &$0.542\pm0.006$& $0.273\pm0.038$\\
        IBAT \cite{IBAT2011}  &  -  & -  & -  &  -  & $0.478\pm0.003$ & $0.075\pm0.001$  & $0.525\pm0.006$ & $0.182\pm0.004$ & $0.486\pm0.006$ & $0.184\pm0.003$\\
        GMVSAE \cite{onlineGMVSAE2020}  & -   &  - &  -  &  -  & $0.512\pm0.015$ & $0.089\pm0.010$ & $0.497\pm0.009$ & $0.170\pm0.005$ & $0.509\pm0.005$ & $0.202\pm0.004$\\
        ATROM  \cite{ATROM}  &  -  & -  &  -  &  -  & $0.499\pm0.002$ & $0.080\pm0.001$ & $0.455\pm0.008$ &  $0.180\pm0.001$ & $0.505\pm0.002$  & $0.191\pm0.002$ \\
       
        SensitiveHUE \cite{Feng2024HUE} &  $0.679\pm0.017$  & $0.143\pm0.047$  & $0.636\pm0.018$   & $0.065\pm0.005$  &  $0.586\pm0.011$  & $0.427\pm0.070$  & $0.698\pm0.007$  &  {\underline{0.756}$\pm0.033$}  &  $0.622\pm0.011$ &  $0.473\pm0.009$\\
        
        
        Transformer \cite{Transformer} &  {\underline{0.804} $\pm~0.005$} &  {\underline{0.162} $\pm~0.013$}  &  {\underline{0.681} $\pm~0.004$} & {\underline{0.263} $\pm~0.009$} &  {\underline{0.761} $\pm~0.011$}  & \underline{0.469}~$\pm0.071$ &  \underline{0.720} $\pm~0.025$ & $0.718\pm0.088$   & \underline{0.695} $\pm~0.004$ & {\underline{0.784} $\pm~0.018$}\\

         \method &   \textbf{0.831} $\pm~0.009$  &  {\textbf{0.230} $\pm~0.042$}  &  {\textbf{0.734} $\pm~0.016$}   &  {\textbf{0.274} $\pm~0.003$}  &   {\textbf{0.817} $\pm~0.009$}  &  {\textbf{1.023} $\pm~0.146$}  &  {\textbf{0.832} $\pm~0.012$}  &  {\textbf{1.877} $\pm~0.178$}  &  {\textbf{0.728} $\pm~0.021$} & {\textbf{0.813} $\pm~0.098$} \\


        \midrule

         Improvement & 3.4\% & 42.0\% & 7.8\% & 4.2\% & 7.4\% & 118.1\% & 15.6\% & 148.3\% & 4.7\% & 3.7\% \\
      
    \bottomrule
    \end{tabular}
    }
    \label{tab:anomaly_detection}
    \vspace{-1em}
\end{table*}

%% file: 05relatedwork.tex

We brief related work here and refer to Appx.~\ref{appendix:other_related_work} for a detailed version and Table~\ref{tab:method_compare} for comparison to the proposed \method.

\noindent \textbf{Human Mobility Modeling.} Methods in this field can be broadly categorized into
traditional statistical and deep learning approaches. Statistical approaches typically rely on specific functional forms such as Poisson/Hawkes processes \citep{Hawkes1971SpectraOS, daley2008point_processes, ogata1998space_time} or Markov Chains \citep{markov1, markov2, markov3, WhereNext} to predict event arrival times/next location, which struggle to capture the intricate spatiotemporal patterns in human mobility. On the other hand, deep learning models based on RNNs \citep{Gao2017,Song2016,du2016recurrent} and Transformer \citep{wan2021pre,DeepMove,MobTCast,Abideen2021taxitrans,wu2020transcrime,want2024trans} show their effectiveness for modeling the complex transition patterns. Transformer-based models have been the de facto most popular models mainly benefiting from their multi-head self-attention for capturing the pair-wise relationship between any pairs in the sequence. For example, CTLE \cite{2021CTLE} learns context and time-aware location embeddings with masked pre-training.  Despite the great success achieved, most of them overlook the underlying uncertainty in the data. This motivates us to develop an uncertainty-aware model to fill the gap, toward more accurate and robust models to modeling human behavior.



\noindent \textbf{Human Mobility Anomaly Detection.} There is limited literature on human mobility anomaly detection; the most related work is trajectory anomaly detection, which aims to judge whether a trajectory is an anomaly given a sequence of GPS points.  
 IBAT \cite{IBAT2011} identifies anomalies by measuring the degree of isolation of the target trajectory from others. 
 GMVSAE \cite{onlineGMVSAE2020} learns trajectory patterns with the Gaussian Mixture Model and uses the likelihood of a target trajectory as the anomaly score. ATROM \cite{ATROM} embeds trajectories using Gaussian priors and uses a threshold-based decision rule to classify them as known or unknown anomalies. 
 

\textbf{Multivariate Time Series (MTS) Anomaly Detection.}  Another related area is MTS anomaly detection \cite{2024DualTF, Feng2024HUE, zamanzadeh2024deep, wang2024revisiting} where each event feature can be treated as a separate channel. SensitiveHUE \cite{Feng2024HUE} proposes a probabilistic network with reconstruction and uncertainty estimation, and uses both terms as anomaly score.
However, neither trajectory nor time series anomaly detection can model the semantic complexity of discrete human event sequences with both numerical and categorical features. 

\textbf{Uncertainty Estimation in Deep Models.} Uncertainty learning in deep models has shown great promise in computer vision \citep{kendall2016,Huang2018semantic} and natural language processing \citep{gal2016,uncertainNLP}. \citet{kendall2017} pioneers the joint modeling of aleatoric and epistemic uncertainty for vision tasks, while ensemble-based approaches \cite{lakshminarayanan2017simple} estimate uncertainties by aggregating predictions.  
Recently, deep evidential models \cite{amini2020deep,sensoy2018evidential,ye2024uncertaintyregularized} and diffusion models \cite{chanestimating,shu2024zero,wen2023diffstg} have been explored and achieved great success. However, deep evidential models typically focus on one feature type while struggling to handle discrete event sequences with both numerical and categorical feature types. Diffusion models generally suffer from complex architectures and time-costing inference.
Besides, none of them are specifically designed $i$) to model human event sequences, nor do they simultaneously $ii$) address anomaly detection.






%% file: 06conclusion.tex
We introduced \method~for uncertainty-aware human mobility modeling and unsupervised anomaly detection. \method integrates both data and model uncertainty estimation, addressing the inherent stochasticity of human behavior and the challenges posed by data scarcity in capturing complex patterns. It enables robust training against noisy input, allows prediction with rejection, and fosters effective anomaly scoring. Experiments on industry-scale data at Novateur showed that \method improves both forecasting and anomaly detection performances over existing baselines. 




%% file: 07appendix.tex
\section{Broader Impact} \label{appendix:broader_impact}
We propose an unsupervised approach that can model uncertainty in sequence data, and provide a robust way to detect anomalies without relying on labeled data. While \method is primarily driven by human mobility data, it is highly generalizable and can be applied across various domains to learn user behavior-related patterns and for anomaly detection. To give some examples, we envision the following scenarios in which our method can be applied:
\begin{itemize}[leftmargin=*]
    \item {Finance}: Detecting anomalies in financial transactions, such as credit card fraud or money laundering, by analyzing sequential spending and money transfer behaviors.
    \item E-commerce and online shopping: Identifying unusual patterns in user purchasing behavior sequences, such as detecting bot-like activity or fraudulent transactions on online platforms.
    \item App user behavior analysis:  Monitoring app user behavior to detect shifts, irregular usage patterns, or engagement anomalies,  providing insights into user retention and app performance.
\end{itemize}
By incorporating uncertainty estimation, \method provides a robust and accurate approach to modeling user behavior and detecting anomalies across these and other application areas.


\section{Model Implementation Details} \label{appendix:model_implementation}

\subsection{Details of Transformer Block} \label{appendix:transformer}
\par \textbf{Transformer Block}, which contains two key components: a Multi-Head Self-Attention (MSA) layer and a Feed-Forward Network (FFN) layer. The MHA layer facilitates message passing between input tokens, while the FFN applies non-linear transformations to enhance feature extraction across different dimensions of the input vectors. To capture more complex interactions between tokens, multi-head attention is employed, where the attention mechanism is defined in Eq.~ \eqref{eq:attention}.
\begin{equation}
{\rm{Attention}}(\bm Q, \bm K, \bm V) = {\mathop{\rm softmax}\nolimits} \left( {\frac{{\bm{Q}{\bm{K}^T}}}{{\sqrt d }}} \right)\bm{V}, 
\label{eq:attention}
\end{equation}
where $\bm{Q} \in \R ^{N \times D}$, $\bm{K} \in \R ^{N \times D}$, and $\bm{V} \in \R ^{N \times D}$ represent the query, key, and value matrices, respectively, all projected from the same input matrix $\bm E$ (which have different forms in Feature-level Transformer and Event-level Transformer) with different learnable weight matrix. The softmax function transforms the scaled dot product into attention weights for $\bm{V}$, and $d$ is the dimensionality of $\bm{K}$ used for scaling the inner product.  Besides, more high-order mutual information can be captured by stacking multiple Transformer blocks. 
Denote the embedding outputted by block $m \in \{ 1, \cdots, M\}$ as $e_i^m$, its updating process can be formulated as follows
\begin{equation}
\begin{array}{l}
{{{{\hat e}}}_i} =  {{{e}}_i^{m - 1} + {\rm{MSA}}_i^m\left( {{\rm{head}}_1^{m - 1}, \ldots ,{\rm{head}}_{n_{\rm{head}}}^{m - 1}} \right)} \\
{{e}}_i^m = {{{{\hat e}}}_i} + {\rm{FFN(}}{{{{\hat e}}}_i}{\rm{)}}.
\end{array}  
\end{equation}
where $n_{\rm{head}}$ is the number of heads.

\subsection{Details of Feature Tokenizer}
\label{appendix:feature_tokenizer}

The feature tokenizer transforms all features ${\bm x} \in {\mathbb R}^{F} $ of an event into embeddings ${\bm e} \in {\mathbb R}^{F \times D}$, where $F$ is the number of input features and $D$ is the embedding dimension.  Specifically,  a numerical feature $x^{(num)}_j$ is projected by a linear transformation with weight {${\bm W}^{(num)}_j \in \R^{D}$} and bias ${\bm b}_j \in \R^{D}$, and the embedding of a categorical feature is implemented as the embedding lookup table ${\bm W}^{(cat)}_j \in \R^{C_{j} \times D}$, where $C_j$ is the total number of categories for feature $x_j^{(cat)}$. Overall:
\begin{alignat*}{2}
    &\e^{(num)}_j  = {\bm b}^{(num)}_j + x^{(num)}_j \cdot {\bm W}^{(num)}_j && \in \R^D \;, \\
    &\e^{(cat)}_j  = {x}^{(cat)}_j {\bm W}^{(cat)}_j             && \in \R^D \;, \\
    &\e            = \mathtt{stack} \left[ \e^{(num)}_1,\ \ldots,\ \e^{(num)}_{k^{(num)}},\ \e^{(cat)}_1,\ \ldots,\ \e^{(cat)}_{k^{(cat)}} \right]              && \in \R^{F \times D} \;.
\end{alignat*}
where ${x}^{(cat)}_j$ is represented as a one-hot vector. The feature embedding ${\bm e} \in \R^{F \times D}$ of one event is the concatenation of all numerical embeddings and categorical embeddings.

\subsection{Details of Time Encoding} \label{appendix:time_encoding}

To encode time-of-day information, we use a cyclical representation by projecting each timestamp onto a two-dimensional circle. Specifically, for each timestamp $t$ (measured in minutes since midnight of each day), we normalize it to the interval $[0, 2\pi]$ and compute the corresponding Cartesian coordinates on a circle of fixed radius $r$:

\begin{equation}
\theta = \frac{t}{1440} \cdot 2\pi, \quad 
\text{time}_x = r \cdot \cos(\theta), \quad 
\text{time}_y = r \cdot \sin(\theta)
\end{equation}
The resulting features $\text{time}_x$ and $\text{time}_y$ are treated as two separate continuous inputs during model training.

At inference time, the model outputs predictions $\hat{\text{time}}_x$ and $\hat{\text{time}}_y$. We then recover the predicted time angle as the angular distance from midnight on a unit circle, where the direction pointing to midnight (i.e., 00:00) is treated as the reference (zero angle) on the circle. The angle is computed as:

\begin{equation}
\hat{\theta} = \texttt{atan2}(\hat{\text{time}}_y, \hat{\text{time}}_x)
\end{equation}

\paragraph{Aleatoric Uncertainty (AU)} The aleatoric uncertainty of the time angle is computed as the average of the AU of $\text{time}_x$ and $\text{time}_y$:

\begin{equation}
\beta_{\text{time angle}} = \frac{\beta_{\text{time}_x} + \beta_{\text{time}_y}}{2}
\end{equation}

\paragraph{Epistemic Uncertainty (EU)} 

To compute epistemic uncertainty for the time angle, we compute the variance of predicted time angles across $T$ forward passes. Let $\hat{\theta}^{(i)}$ denote the time angle predicted in the $i$-th pass. We first compute the mean predicted angle:

\begin{equation}
\bar{\theta} = \texttt{atan2}\left( \frac{1}{T} \sum_{i=1}^{T} \hat{\text{time}}_y^{(i)}, \; \frac{1}{T} \sum_{i=1}^{T} \hat{\text{time}}_x^{(i)} \right)
\end{equation}

Next, we compute the angular deviation $\delta_i$ of each prediction from the mean, taking the shorter arc on the circle:
\begin{equation}
\delta_i = \min\left(|\hat{\theta}^{(i)} - \bar{\theta}|, \; 2\pi - |\hat{\theta}^{(i)} - \bar{\theta}|\right)
\end{equation}

The epistemic uncertainty is the mean squared angular deviation:

\begin{equation}
\alpha_{\theta} = \frac{1}{T} \sum_{i=1}^{T} \delta_i^2
\end{equation}

\section{Using Uncertainty  for Anomaly Scoring}
\label{appendix:uncertainty}

\subsection{Relationship between Loss and EU}
\label{appendix:loss_eu}
\begin{figure}[htbp]
    \centering
    \includegraphics[width=0.9\linewidth]{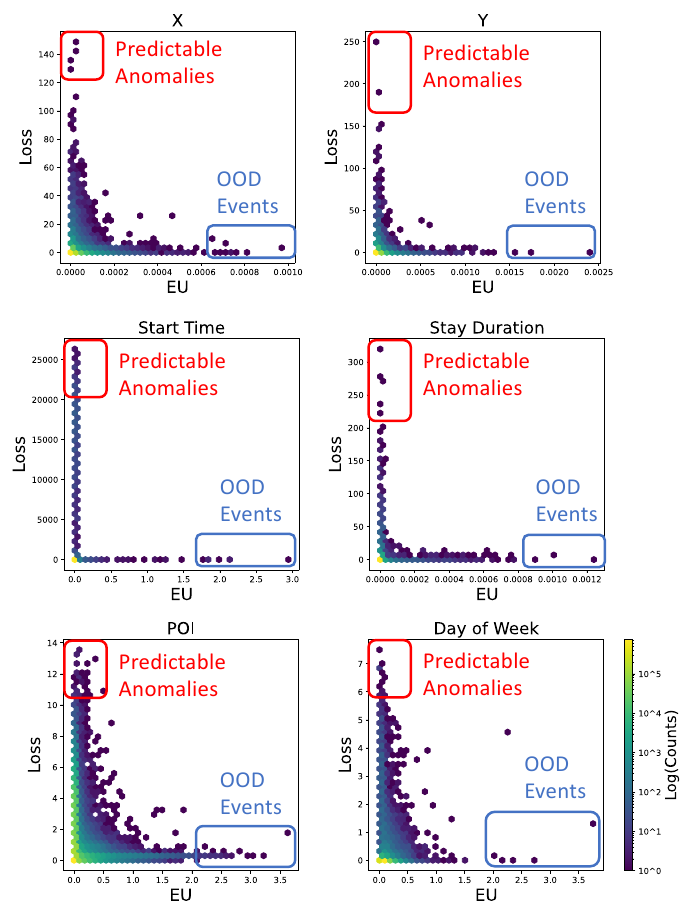}
    \caption{Relationship between prediction loss and EU for all features. Color denotes the logarithm of the count of events in each hexagon. Region for predictable anomalies and high-EU events can be observed for all features.}
    \label{fig:loss_eu_all_features}
\end{figure}

We illustrate the Relationship between prediction loss and EU for all features in \red{Figure~\ref{fig:loss_eu_all_features}}.

\subsection{Relationship between EU and kNN Distance}
\label{appendix:eu_knn}

\begin{figure}[htbp]
    \centering
    \includegraphics[width=0.6\linewidth]{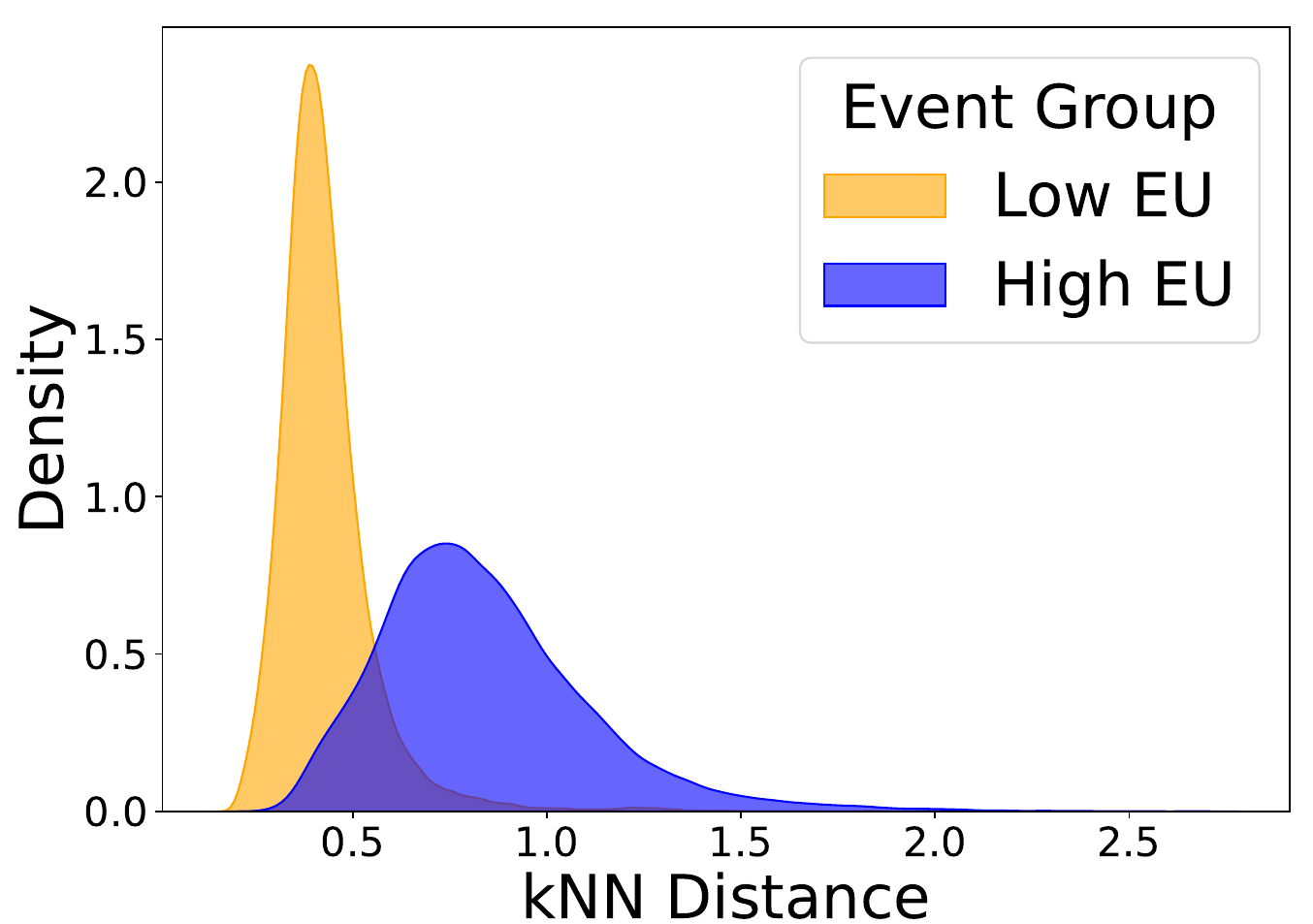}
    \caption{Distribution of kNN distance of test events to training events in the embedding space for high-EU and low-EU events. High-EU and low-EU events are defined as the top and bottom 5\% of events ranked by their EU, respectively. We select k to be 150 to balance between capturing local clusters and accounting for outlier distances. Events with very high EU often exhibit large nearest-neighbor distances in the embedding space of our \method.}
    \label{fig:eu_knn_hist}
\end{figure}

\begin{figure}[htbp]
    \centering
    \includegraphics[width=0.8\linewidth]{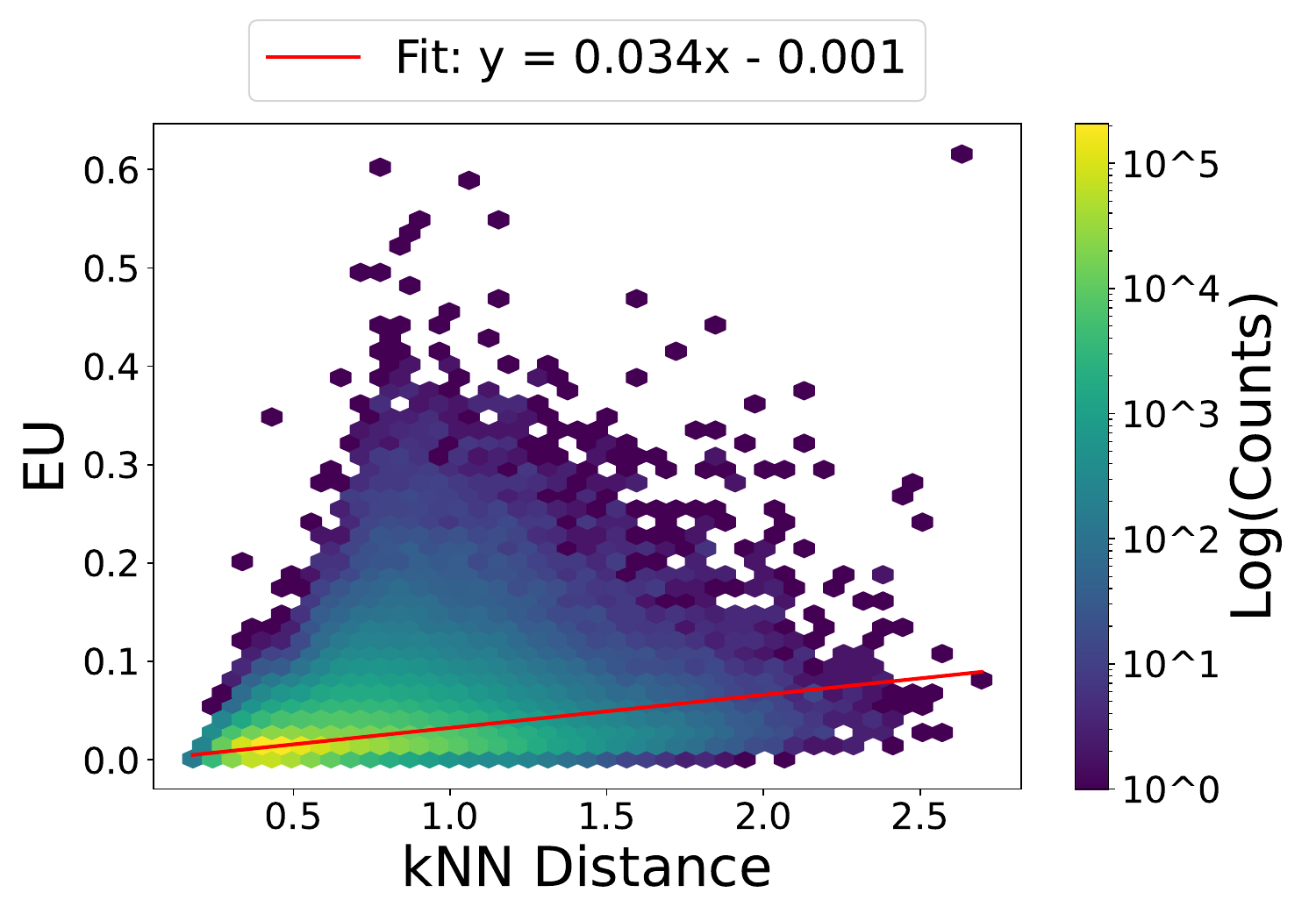}
    \vspace{-1em}
    \caption{Hexbin plot of kNN distance of test events against EU. Color denotes the logarithm of the count of events in each hexagon. The linear fit shows a weak association between OOD-ness (as measured by kNN distance) and EU.}
    \label{fig:eu_knn_scatter}
    \vspace{-0.1in}
\end{figure}

In \red{Figure~\ref{fig:eu_knn_hist}}, we observe that compared to low-EU events, high-EU events tend to exhibit larger kNN distances. However, as shown in \red{Figure~\ref{fig:eu_knn_scatter}}, the overall relationship between EU and kNN distance across all events is weak. This finding indicates that while very high EU values may correspond to events that are outliers in the embedding space, EU alone is not a reliable indicator of the out-of-distribution events for the broader dataset.

\subsection{Experiments on Self-injected Anomalies}
\label{appendix:injection}
We design two categories of injected anomalies to mimic predictable anomalies and out-of-distribution (OOD) anomalies. 

\textbf{Predictable anomalies} Predictable anomalies are created by altering individual events' spatial or temporal features for randomly selected agents. (1) Spatial Alteration: The POI of a randomly selected event is changed to one of the 20 least visited POIs. A new location within this POI category is then randomly assigned to the event. (2) Temporal Alteration: Noise is added to the event's start time and stay duration. We hypothesize that these feature alterations should result in increased prediction loss. 

\textbf{OOD anomalies} OOD anomalies are designed to introduce patterns significantly different from those in the training data: (4) Swap: For two randomly selected agents, one day of events is swapped between them. (5) Permutation: A six-day window is randomly selected from an agent’s event sequence. While the temporal features remain unchanged, the spatial features are permuted to create a new sequence. (6) High Density: A three-day window is randomly selected from an agent’s event sequence. The original events are replaced with dense activity patterns, where the agent visits many randomly selected locations within short durations.

\begin{figure}[htbp]
  \includegraphics[width= 1\linewidth]{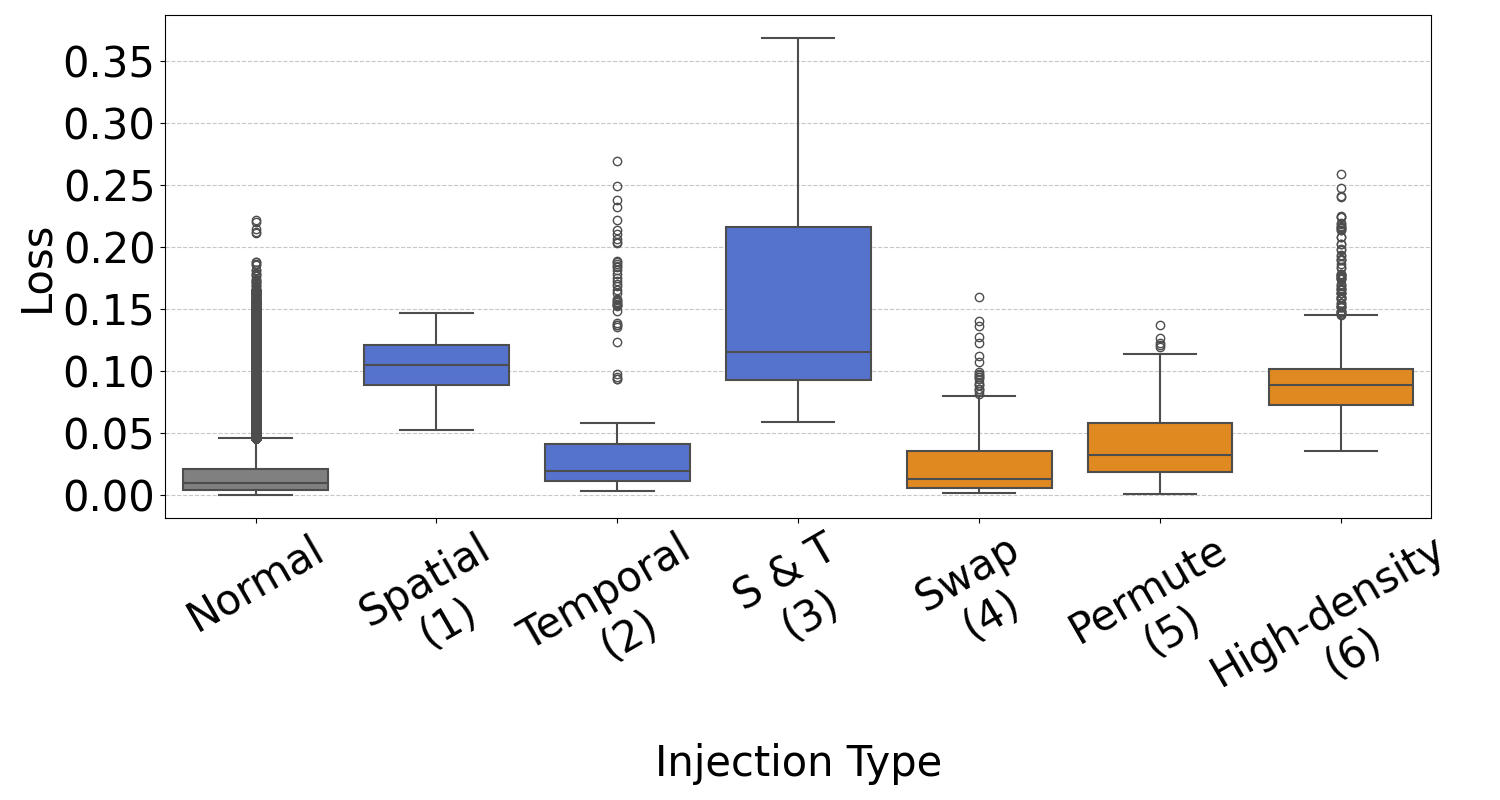}
   \vspace{-1.5em}
  \caption{ Boxplot of loss for normal events and different types of injected anomalies. Event loss is aggregated via normalizing feature loss to $[0,1]$ and taking the average. }
  \label{fig:loss_injection}
\end{figure}

\begin{figure}[htbp]
  \includegraphics[width= 1\linewidth]{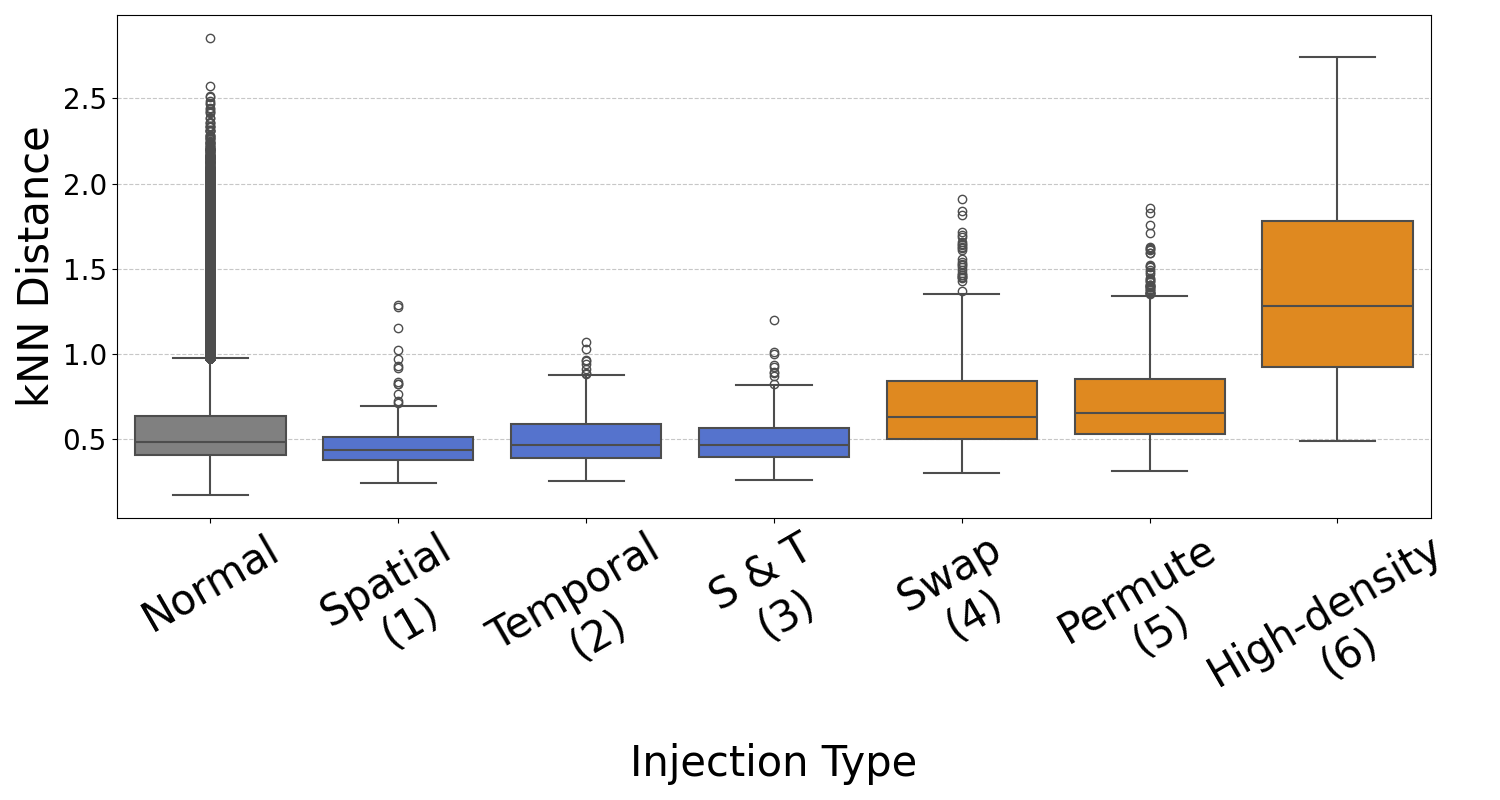}
   \vspace{-1.5em}
  \caption{Boxplot of kNN distance for normal events and different types of injected anomalies. S\&T denotes that both spatial and temporal features are altered.}
  \label{fig:knn_injection}
\end{figure}

We inject the anomalies into \trial and conduct the following analysis and experiments.
An analysis of the loss distribution (see \red{Figure~\ref{fig:loss_injection}}) reveals that the predictable anomalies show a higher loss compared to normal events and OOD anomalies. 
To evaluate the effectiveness of our anomaly detection methods, we first test the performance of using prediction loss and EU individually on the injected anomalies. As shown in \red{Table~\ref{tab:ablation_injection_1}}, EU alone fails to effectively capture OOD anomalies. Next, we analyze the distribution of kNN distances for different anomaly types (see \red{Figure~\ref{fig:knn_injection}}). We observe that OOD anomalies exhibit higher kNN distances compared to predictable anomalies, indicating that kNN distance effectively distinguishes between these two categories. Based on these findings, we incorporate kNN distance into the anomaly scoring. \red{Table~\ref{tab:ablation_injection_1}} shows that kNN distance, combined with prediction loss, achieves better detection performance particularly when the injected anomalies are diverse.

\input{tables/tab_injection_AD_1}


\section{Data Details}
\label{appendix:data_details}

\subsection{Data Statistics} \label{appendix:data_stat}
The statistics of the dataset are illustrated in \red{Table~\ref{tab:stat_test_data}}.
\input{tables/tab_dataset}

\subsection{Details on Data Simulation \& Anomaly Injection} \label{appendix:data_simulation}

\textbf{Why Simulated Data?} There are several reasons for conducting experiments on the simulated data: (1) First and most importantly, to the best of our knowledge, there is currently no human activity data available at the industry scale with anomaly labels, except for \NUMOSIM. Though there are many trajectory datasets adopted by previous works, they do not include anomaly labels. (2) Manually collecting human trajectories and labeling anomalies is a time-consuming and resource-intensive process, requiring significant effort and expertise. 
(3) Simulated data enables domain experts, such as "red teams," to inject various types of realistic anomalies systematically. This allows for a comprehensive evaluation of the model's ability to detect diverse anomaly types.

\textbf{Details on Simulation System.} Motivated by the above observation, we next introduce the system in Novateur for mobility data generation and anomaly detection. Overall, the simulation system is built with three goals: $i)$ \textit{realistic}, i.e., all simulated GPS is based on a real-world city and human behaviors are aligned with the real world (such as staying home in the evening and going to work in the morning); $ii)$ \textit{complex}, the generated data should contain diverse agents with different behavior patterns (which we do not know); $iii)$ \textit{large-scale}, the generated data should contain an extensive number of agents and events to facilitate the training and evaluation of different models.

Specifically, the datasets are generated using Deep Activity Model (DeepAM) \cite{liao2024deepactivity}, an advanced generative deep learning model trained on 180,000 samples from the 2017 National Household Travel Survey (NHTS) \cite{fhwa2017nhts}. As the authoritative source on American travel behavior, NHTS provides sociodemographic data of individuals and households, along with their travel patterns, enabling DeepAM to simulate a diverse range of realistic mobility scenarios, including both typical and anomalous behaviors. The simulation follows a three-step process: (1) Activity Sequence Generation: DeepAM generates daily activity sequences with start and end times for each agent. (2) Activity-to-POI Mapping: Each generated activity is assigned to a real-world Point of Interest (POI) based on data from PlanetSense \cite{ornl2024planetsense} and USA Structures \cite{fema2024usastructures}. (3) Mobility Simulation: Agents navigate the real road network (sourced from OpenStreetMap \cite{openstreetmap}) to move between POIs, ensuring realistic travel patterns.

To mimic realistic behavior anomalies, the datasets introduce two types of anomalies: non-recurring and recurring. (1) Non-recurring anomalies are one-time deviations from an agent’s usual behavior during the test period. For example, the agent visits a new, unexpected location instead of a usual POI, or leaves a stay point significantly earlier or later than expected. (2) Recurring anomalies represent systematic behavioral shifts over time, simulating longer-term deviations from established patterns. For example, the agent regularly visits a new location (e.g., dining at a new restaurant every weekend instead of the usual one), or there are changes in visit frequency, timing, or sequence of activities. Further details can be found in \cite{numosim}. We use 3 datasets from the simulation (statistics in \red{Table~\ref{tab:stat_test_data}} which covers cities of different scales.

\section{Details on Masked Prediction}\label{appendix:masked_prediction_detail}

\subsection{Baselines} \label{appendix:masked_prediction_baselines}
\par The following methods are chosen as baselines. Since the masked prediction is not the focus of this work, therefore we choose and implement several backbones (proven to be effective in many previous works) in many human mobility models as baselines:
\begin{itemize}[leftmargin=*]
    \item MLP \citep{MLP}, which applies multiple Linear layers to encode the event sequence, and decodes the masked features according to the embedding from the last layer.
    \item LSTM \citep{LSTM},  a recurrent neural network commonly used to model sequential data, reads the event sequence and decodes the masked event based on its hidden vector.
    \item Transformer \citep{Transformer}, one of the most powerful sequence models nowadays, which models the correlation of different events by multi-head self-attention.
  
    \item LightPath \cite{2023LightPath} is a self-supervised model for human trajectory representation learning. To train the model, it masks each sequence with two mask ratios (we use 0.05 and 0.1) to get different views of the raw sequence. The loss has two components: (1) the reconstruction loss that recovers each masked sequence; and (2) the contrastive loss (binary classification) that treats two representations from the same raw sequence as positive pairs and others in the mini-batch as negative pairs. Note that LightPath does not support the classification task, so we do not report the POI prediction performance in the experiments. 

    \item Dual-Transformer, a variant of our proposed model, which removes the uncertainty learning mechanism and only keeps the dual Transformer architecture. 

\end{itemize}


\subsection{Settings} \label{appendix:masked_prediction_settings}
\par \textit{Metrics.} We consider the prediction of numerical features and categorical features as regression and classification tasks, respectively. To this end, we apply MAE and MAPE to evaluate the performance of regression, and Accuracy (ACC) is applied to evaluate the classification performance. The units for x and y are kilometers while the units for start time and stay duration are minutes.

\par \textit{Settings.}  The models are trained on 1 GPU of NVIDIA RTX A6000.  For all deep models, we search the parameters in a given parameter space, and each model's best performance is reported in the experiment. For the proposed \method, we introduce dropout layers after each feature's embedding layer with a dropout ratio of 0.05, which randomly drops a subset of neurons during both training and inference. The batch size of each epoch is 128 and the learning rate of Adam optimizer starts from 1e-3 with a weight decay of 1e-05. The parameter search space of \method is: the mask ratio for pre-training is searched from [0.05, 0.3]. The embedding size $D$ for the transformer is searched from [32, 64, 128].  The layer $M_1$ for the feature-level transformer is set to 1 for simplicity, and the layers $M_2$ for the event-level transformer are searched from [3, 4, 5].  For CTLE \cite{2021CTLE}, we use the default parameters in the paper, which are 200 epochs for pertaining, and 128 for embedding size.

\subsection{Masked Prediction Results} \label{appendix:masked_prediction_full_results}

The masked prediction results in \trialthree can be found in \red{Table~\ref{tab:masked_prediction_trial3}}.

\input{tables/tab_masked_prediction_trial3}

\section{Details for Anomaly Detection}
\label{appendix:anomaly_detection_details}

\subsection{Baselines} \label{appendix:anomaly_detection_baselines}
There are only a few related public achievements for human mobility anomaly detection, thus it is difficult to find state-of-the-art models for direct comparison. For a comprehensive evaluation, we select baselines from three domains for comparison: 
\par  (1) Human mobility Learning:
\begin{itemize}[leftmargin=*]
    \item CTLE \cite{2021CTLE} learns context and time-aware location embeddings. 
\end{itemize}


\par (2) Trajectory anomaly detection: 
\begin{itemize}[leftmargin=*]
    \item IBAT \cite{IBAT2011} detects anomalies using how much the target trajectory can be isolated from other trajectories.
    \item  GMVSAE \cite{onlineGMVSAE2020} uses a generative variational sequence autoencoder model that learns trajectory patterns with Gaussian Mixture model and uses the probability of trajectory being generated as the anomaly score.
    \item  ATROM \cite{ATROM} uses variational Bayesian methods and correlates trajectories with possible anomalous patterns with the probabilistic metric rule.
\end{itemize}

\par (3) Multi-variate time series anomaly detection: 
\begin{itemize}[leftmargin=*]
    \item SensitiveHUE \cite{Feng2024HUE} uses a probabilistic network by implementing both reconstruction and heteroscedastic uncertainty estimation, and uses both terms as anomaly score.
\end{itemize}

\par Method currently adopted by Novateur:
\begin{itemize}[leftmargin=*]
    \item  \ssmlp, which takes event features including GPS coordinates, start time, stay duration, and POI type as input. These features are first converted into embeddings, which are then concatenated to form a unified input vector. This vector is processed by an 8-layer MLP that outputs a probability indicating whether an event is real (1) or fake (0). During training, fake events are generated by randomly altering features of real events, and the model learns to classify real and fake using binary cross-entropy loss. This approach allows the model to distinguish between normal and anomalous data. Finally, the negative of the output probability is used as the anomaly score, with a higher score indicating greater anomaly likelihood. 
\end{itemize}

\subsection{Settings} \label{appendix:anomaly_detection_setting}

Next, we describe how each baseline method computes anomaly scores and the hyperparameters used in our experiments. 
\begin{itemize}[leftmargin=*]
    \item For Transformer, we compute prediction error for numerical features as $\Delta_f^{\text{num}}= |\hat{y}_f - y_f|$ and for categorical features as $\Delta_f^{\text{cls}}= 1 - p_{f,c}$ where $c$ is the true class. 

    \item For CTLE \cite{2021CTLE}, since the method is most effective for dense trajectories, we first convert GPS coordinates into grids and use grid IDs as location identifiers. We then adapt the original implementation to incorporate stay duration, latitude, longitude, and POI type, alongside location and time, for learning location embeddings. Next, we train a transformer to reconstruct event features from these learned embeddings. Finally, we apply the same anomaly scoring approach as used for Transformer.  The embedding size is searched from [64, 128].
    \item IBAT \cite{IBAT2011} determines anomaly scores based on how easily a trajectory can be isolated from others, where easier isolation indicates a rarer pattern and higher anomalousness.  The number of running trials is searched from [50, 100, 200, 300] and the subsample size is searched from [64, 128, 256, 512].

    \item GMVSAE \cite{onlineGMVSAE2020} computes anomaly score as the probability of the trajectory being generated from the learned patterns. The embedding size is searched from [256, 512] and the number of clusters is searched from [5, 10]. 

    \item ATROM \cite{ATROM} assigns anomaly scores based on the probability of a trajectory being classified into predefined anomaly categories. The embedding size is searched from [64, 128, 512].

    \item SensitiveHUE \cite{Feng2024HUE} computes anomaly score for time $t$ and channel $s$ as $S(t,s) = \frac{(\hat{\mu}_{ts} - X_{ts})^2}{2 \hat{\sigma}_{ts}^2} + \frac{1}{2} \ln \hat{\sigma}_{ts}^2$, where $\hat{\mu}_{ts}$ is the reconstruction, $X_{ts}$ is the input, and $\hat{\sigma}_{ts}^2$ is the estimated uncertainty. The final score across multiple channels is $\max_{s} \tilde{S}(t,s) = \frac{S(t,s) - \text{Median}(s)}{\text{IQR}(s)}$. The embedding size is searched from [64, 128] and the number of layers is [1, 2].
\end{itemize}
  
For all baselines, the agent-level anomaly score is defined as the maximum event score across all events for a given agent.



\section{Details on Experiment Results}

\subsection{AU and EU under different among of training data} \label{appendix:AU_EU_different_training_size}


We investigate how AU and EU estimates change with varying amounts of training data, under three configurations with 1/2/3 weeks of training data and 2 weeks of test data on \trial. \red{Table~\ref{appendix:tab:uc_train_data_relation}} reports the average AU and EU across test samples for feature x and POI (for all features. We observe that ($i$) AU remains almost constant as training data size increases, which empirically shows that AU accounts for the inherent randomness/stochasticity in the data that is not a matter of the amount of training data, and ($ii$) EU decreases noticeably with increasing data size, suggesting it can be reduced with sufficient training data.

\input{tables/tab_uc_train_data_relation_appendix}

\subsection{Ablation on Anomaly Scores} \label{appendix:anomaly_scores}

\begin{figure}[htbp]
\vspace{-0.1in}
  \includegraphics[width= 0.95\linewidth]{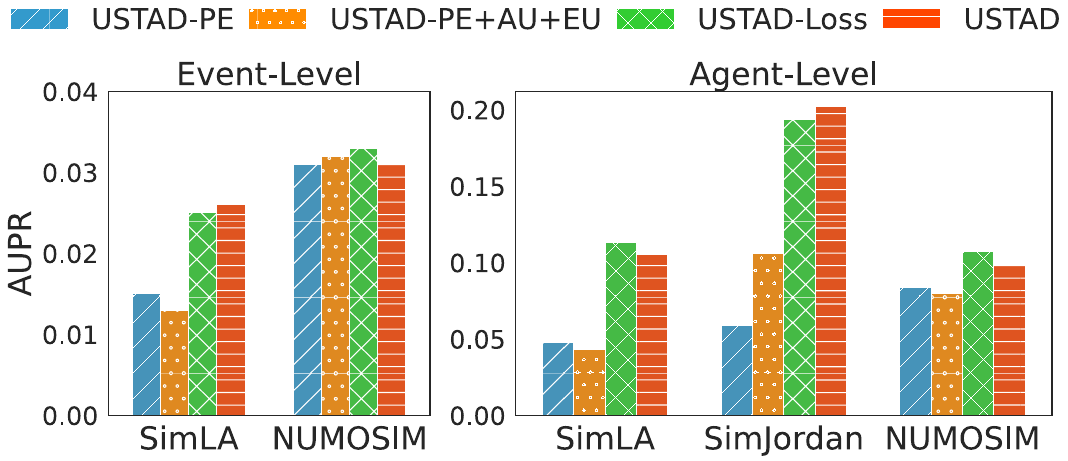}
  \vspace{-0.1in}
  \caption{Ablation on different anomaly scores. PE denotes prediction error. Results demonstrate the effectiveness of incorporating uncertainty estimates and OODness (by kNN) into anomaly scoring.}
  \label{fig:score_ablation_aupr}
\vspace{-0.2in}
\end{figure}

\red{Figure~\ref{fig:score_ablation_aupr}} shows the AUPR performance of different anomaly scoring functions in \trial.

\subsection{More Case studies} \label{appendix:more_cases}


\begin{figure*}[htbp]
    \centering
    \includegraphics[width=1 \linewidth]{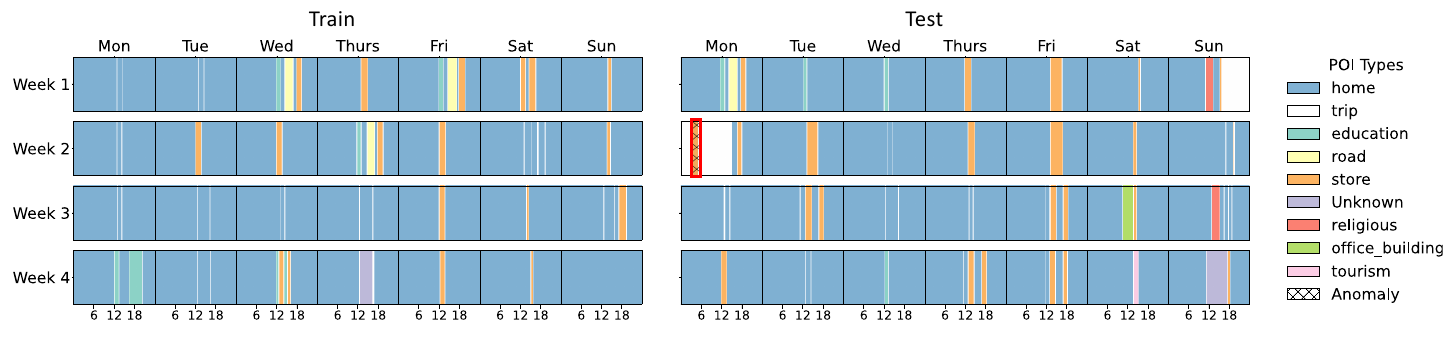}
    \caption{A case of ``predictable anomaly'' (i.e. high-loss \& low-EU)  w.r.t. start time (ST): During the training period (left), the agent consistently stays at home in the early morning and only visits the store around noon or in the early afternoon. However, in the test period (right), the agent makes an unusual early-morning visit to a store. Hence, the model confidently predicts (low EU) the typical visit times (i.e. noon or early afternoon) not the observed time (i.e. early morning) to store, resulting in a high loss for the start time.}
    \label{fig:high_time_loss}
\end{figure*}
\begin{figure*}[htbp]
    \centering
    \includegraphics[width=1 \linewidth]{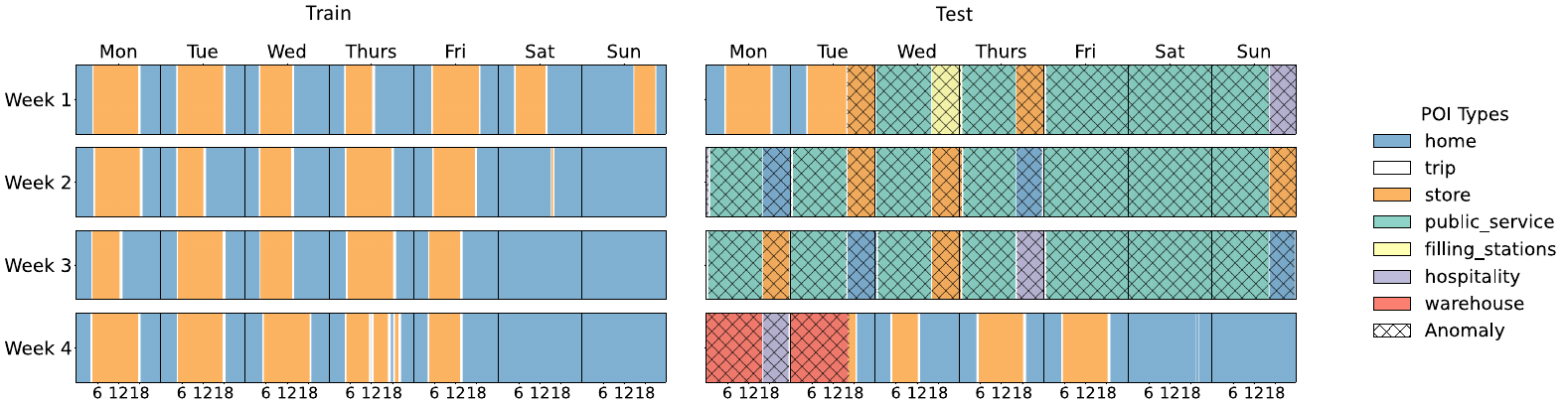}
    \caption{A case of high-kNN anomaly event: The agent consistently visits only home and stores with regular patterns during training (left) but in test (right) s/he recurringly visits new POIs—such as filling stations, hospitality facilities, and warehouses—at unusual times (evening or late night) and for long durations. These events at previously unseen POIs combined with atypical timing and duration, result in large kNN distances from the training events in the embedding space.}
    \label{fig:high_knn}
\end{figure*}
\begin{figure*}[htbp]
    \centering
    \includegraphics[width=1 \linewidth]{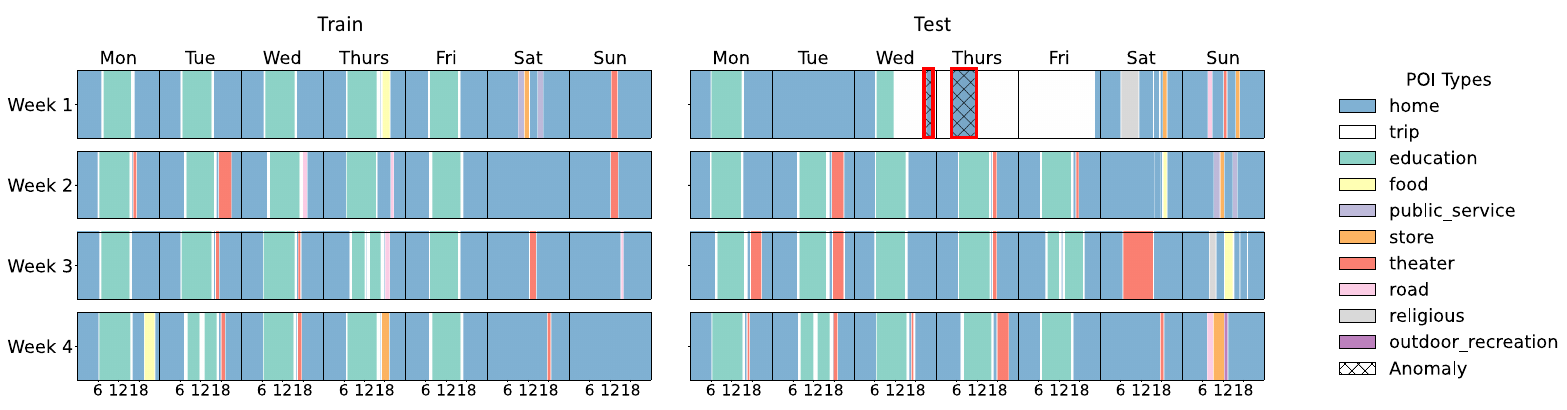}
    \caption{A case of high-kNN anomaly event: The agent consistently visits home and education facilities with regular patterns, occasionally other POI with short durations, during training (left). However, on one day in the test period (right), s/he leaves home late at night, returns to it early morning, leaves again, and takes a long trip before returning. This behavior is very different from her/his routine, resulting in large kNN distances from the training events.}
    \label{fig:high_knn_2}
\end{figure*}

We present more cases in this section. \red{Figure~\ref{fig:high_time_loss}} illustrates one case of High-loss \& low-EU for the feature ``start time''. During training (left), the agent consistently stays at home in the early morning. However, in the test period (right), the agent makes an unusual early-morning visit to a store. Since the store is a previously visited POI, the model confidently (low EU) predicts the typical times but the actual start time is very different (high loss).
In contrast, as we show in  \red{Figure~\ref{fig:high_knn}}, high kNN distance identifies recurring anomalies, where an agent exhibits persistent yet unusual transitions between events or repeatedly visits unseen POIs. Since these behaviors do not resemble any patterns in the training data, they stand out in the embedding space, leading to large kNN distances. Another high-kNN anomaly event (see \red{Figure \ref{fig:high_knn_2}}) occurs when, during the test period, the agent takes a day-long trip that differs significantly from their routine in the training period, resulting in a large kNN distance.
By combining both loss and kNN distance, \method~is effective in capturing both one-time deviations and systematic behavioral shifts, ensuring robust anomaly detection across different scenarios. 

\section{Other Related Work} \label{appendix:other_related_work}

\par We give a detailed introduction of related works in this section.

\textbf{Human Mobility Modeling.} Methods in this field can be broadly categorized into
traditional statistical and deep learning approaches. Statistical approaches typically rely on specific functional forms such as Poisson/Hawkes processes \citep{Hawkes1971SpectraOS, daley2008point_processes, ogata1998space_time} or Markov Chains \citep{markov1, markov2, markov3, WhereNext} to predict event arrival times/next location, which struggle to capture the intricate spatiotemporal patterns in human mobility. On the other hand, deep learning models based on RNNs \citep{Gao2017,Song2016,du2016recurrent} and Transformer \citep{wan2021pre,DeepMove,MobTCast,Abideen2021taxitrans,wu2020transcrime,want2024trans} show their effectiveness for modeling the complex transition patterns. Transformer-based models have been the de facto most popular models mainly benefiting from their multi-head self-attention for capturing the pair-wise relationship between any pairs in the sequence. For example, CTLE \cite{2021CTLE} learns context and time-aware location embeddings with masked pre-training.  Despite the great success achieved, most of them overlook the underlying uncertainty in the data. This motivates us to develop an uncertainty-aware model to fill the gap, toward more accurate and robust models to modeling human behavior.


\noindent \textbf{Human Mobility Anomaly Detection.} There is limited literature on human mobility anomaly detection; the most related work is trajectory anomaly detection, which aims to judge whether a trajectory is an anomaly given a sequence of GPS points. For example, IBAT \cite{IBAT2011} detects anomalies using how much the target trajectory can be isolated from other trajectories. GMVSAE \cite{onlineGMVSAE2020} uses a generative variational sequence autoencoder model that learns trajectory patterns with the Gaussian Mixture model and uses the probability of trajectory being generated as the anomaly score. ATROM \cite{ATROM} uses variational Bayesian methods and correlates trajectories with possible anomalous patterns with the probabilistic metric rule. However, since trajectory anomaly methods are primarily proposed for uniformly sampled GPS points, thus lack the ability to model the semantic complexity of human activities. Besides, such models cannot perform event-level anomaly detection when applied to our problem. 

\textbf{Multivariate Time Series Anomaly Detection}
Another related effort is multivariate time series (MTS) anomaly detection, if we take one event feature as a feature channel in MTS. SensitiveHUE \cite{Feng2024HUE} proposes a probabilistic network with both reconstruction and heteroscedastic uncertainty estimation, and uses both terms as anomaly score.
However, compared with our model, time series anomaly detection cannot process the discrete human event sequences with both numerical and categorical features.

\noindent \textbf{Uncertainty Estimation.}  Uncertainty quantification has long been studied in probability and statistics. Bayesian inference computes uncertainty through posterior distributions over parameters \citep{gelman2013bayesian, 1998Bayes}, while Gaussian Processes (GP) model uncertainty in regression, particularly in non-parametric settings \citep{rasmussen2006gaussian}. In frequentist statistics, uncertainty is typically measured using confidence intervals and hypothesis testing \citep{wasserman2004all}. Beyond statistics, uncertainty estimation is also important in engineering and robotics, particularly in decision-making, control, and policy learning. For example, in robotics, \citet{2015gaussianrobotics} use GP models to support motion planning and reinforcement learning under uncertainty. In control theory, uncertainty-aware models \citep{Schneider1997control} enhance system stability and robustness in dynamic environments.

Recently, there has been a rising research interest in the uncertainty of deep models. Such efforts in computer vision \citep{kendall2016,Huang2018semantic} and natural language processing \citep{gal2016,uncertainNLP} show great promise of uncertainty learning for task performance improvement. \citet{kendall2017} propose to learn aleatoric uncertainty through loss attenuation while modeling epistemic uncertainty using MC Dropout in both regression and classification tasks for CV. \citet{dropconnect} drop individual weights of the network during training and inference to measure uncertainty robustly. \citet{lakshminarayanan2017simple} employs deep ensembles for model uncertainty estimation. Moreover, deep evidential models \cite{amini2020deep,sensoy2018evidential,ye2024uncertaintyregularized} and diffusion models \cite{chanestimating,shu2024zero,wen2023diffstg}  are proposed to model the uncertainty. However, deep evidential models typically focus on one feature type while struggling to handle discrete event sequences with both numerical and categorical feature types. Diffusion models generally suffer from complex architectures and time-costing inference.


More recently, uncertainty estimation has become a crucial topic in large language models (LLMs) and generative AI, particularly in addressing hallucinations.
One stream of work involves unsupervised methods that quantify uncertainty using the entropy of the token-level predictive posterior distribution \citep{Malinin2021UncertaintyEI, xiao2021hallucination, zhang2023enhancin}. Another stream leverages semantic entropy, which clusters generated outputs into semantically equivalent groups and aggregates their likelihoods to produce an uncertainty measure \citep{kuhn2023semanticuncertainty, duan2024relevance}. Additionally, some works explore prompting LLMs to self-assess their uncertainty by explicitly generating confidence scores or uncertainty estimates about their responses \citep{lin2022teaching, groot2024verbalize}. 
Although the above-mentioned methods have achieved great success, none of these are i) for event sequences or human mobility, nor are they for ii) anomaly detection, which inspires us to propose \method for human mobility anomaly detection.

%% file: tables/tab_injection_AD_1.tex
\begin{table}
    \caption{Anomaly detection results at event level using different scoring functions for \trial with different types of injected anomalies.}
    \vspace{-0.1in}
    \centering
    \renewcommand\arraystretch{1.1}
    \setlength\tabcolsep{2 pt}
    \resizebox{1\linewidth}{!}{
    \begin{tabular}{c|cc|cc|cc|cc}
    \toprule
        \multirow{2}{*}{Injection Type} & \multicolumn{2}{c|}{Loss} & \multicolumn{2}{c|}{EU} & \multicolumn{2}{c|}{kNN} & \multicolumn{2}{c}{Loss \& kNN} \\
        \cline{2-9}
          & AUROC & AUPR & AUROC & AUPR & AUROC & AUPR  & AUROC & AUPR\\
         \midrule
         Spatial  &  0.996 & 0.146 & 0.509  &  0.0002 & 0.378 &0.0001 & 0.996 &  0.135\\
         Temporal & 0.933 & 0.137 & 0.517  & 0.0002  & 0.452 &0.0001&0.926 & 0.126\\
         Spatial \& Temporal& 0.997 & 0.266 & 0.519 & 0.0002 &0.441& 0.0001 &0.997 & 0.251 \\
         Swap & 0.957 &0.152  & 0.608  & 0.001  & 0.692  & 0.001  &0.957 & 0.141 \\
         Permutation & 0.768 & 0.020 & 0.615 & 0.004  & 0.726 & 0.005 &0.771 & 0.018 \\
         High-density & 0.986 & 0.175 & 0.780 &0.017 &0.955 &0.268 & 0.989 & 0.260\\
         All types & 0.899 & 0.254 & 0.676  & 0.017  &0.794 & 0.131  & 0.902 & 0.292\\
    \bottomrule
    \end{tabular}
    }
    \label{tab:ablation_injection_1}
    \vspace{-0.1in}
\end{table}

%% file: tables/tab_dataset.tex
\begin{table}[htbp]%
	\centering
	\caption{Dataset statistics. Note that for \trialthree, the event-level ground truth labels are not provided.}
 \vspace{-0.1in}
	\setlength\tabcolsep{3pt}  
	\resizebox{\linewidth}{!}{  
		\begin{tabular}{lccc}
			\toprule
			& \trial &\trialthree & \NUMOSIM\\
			\midrule
			\#Agents & 20,000 & 20,000 & 20,000 \\
			\#Anomalous Agents & 162 & 370 & 381 \\
                Ratio of Anomalous Agents & 0.008 & 0.019 & 0.019 \\ \midrule
			
			\#Events & 3,111,712 & 3,656,233 & 3,428,714 \\ 
			\#Anomalous Events & 892 & - & 3,468 \\ 
                Ratio of Anomalous Events & 0.0003 & - & 0.001 \\ \midrule
                x range (km) & [-57, 57] & [-11, 11] & [-59, 59]\\
                y range (km) & [-40, 40] & [-10, 10] & [-60, 60]\\
                Avg of stay duration (min) & 509 & 484 & 455 \\
                Avg of start time (min) & 850 & 835 & 776\\ 

                \#{\rm poi} &  39 & 7 & 28 \\

                \midrule

                Time Span & 2 months & 2 months & 2 months \\ 
                City & Los Angeles  & Jordan & Los Angeles \\
                Public/Private & Private  & Private & Public\\

			\bottomrule
		\end{tabular}
	}
	\label{tab:stat_test_data}
\end{table}

%% file: tables/tab_masked_prediction_trial3.tex
\begin{table}[htbp]
\centering
\small
\caption{Masked prediction results. We \textbf{bold} the best performance and \underline{underline} the second best. Abbr: x:lat., y:long. (km), ST: start time (min), SD: stay dur. (min). }
\vspace{-0.1in}
\renewcommand\arraystretch{1.1}
\setlength\tabcolsep{2 pt}
\resizebox{1.0 \linewidth}{!}{
\begin{tabular}{c|cccc|cccc|c}
\toprule 
\multirow{3}{*}{Method}
& \multicolumn{9}{|c}{\trialthree}  \\
\cline{2-10}
& \multicolumn{4}{|c|}{MAE $\;\;\downarrow$}& \multicolumn{4}{|c|}{MAPE (\%) $\;\;\downarrow$}& \multicolumn{1}{|c|}{ACC(\%) $\;\;\uparrow$} \\
\cline{2-10}
 & x & y & ST & SD &  x & y & ST & SD & POI   \\
\midrule 

MLP \citep{MLP}  & 2.62 & 2.14 & 165.25 & 525.02 &212.30 & 266.31& 21.89 & 398.85 & 35.73 \\  

LSTM \citep{LSTM} & 1.31 & 1.18 & 51.15 & 504.19 & 155.21& 168.01 & 19.16 & 463.88  & 51.59\\ 

Transformer \citep{Transformer} & 0.95 & 0.90 & 34.55& 239.35 & 110.37&\underline{128.26}  & \underline{1.52} & 124.69 & 70.42\\ 


Dual-Tr & 0.95 & \underline{0.87}  & 29.33 & 214.33 & 111.39 & 135.30 & 2.09 & 111.11  & 72.72\\ 


LightPath \cite{2023LightPath}  &   1.04  &  0.94  &  47.57  &  274.81  &  128.31  &  144.59  &  \textbf{0.99}  &  197.61  &  -\\ 

\midrule \hline

\method & \underline{0.92} & 0.88 & \underline{26.24} & \underline{142.99} & \underline{97.58} & 138.24 & 2.45 & \underline{43.35} & \underline{75.01} \\

\methodwrej & \textbf{0.66} & \textbf{0.67} & \textbf{18.66} & \textbf{60.44} & \textbf{57.58} & \textbf{98.81} & 1.89 & \textbf{35.76} & \textbf{77.63}   \\

\bottomrule


\end{tabular}
}

\label{tab:masked_prediction_trial3}
\end{table}

%% file: tables/tab_uc_train_data_relation_appendix.tex
\begin{table*}[htbp]
\centering
\caption{Aleatoric (AU) and epistemic uncertainties (EU) per sample under different training data size settings.}
\vspace{-0.1in}
\resizebox{0.95 \textwidth}{!}{%
\begin{tabular}{c|c|c|c|c|c|c|c|c|c|c|c}
\toprule
Train & Test & \multicolumn{2}{c|}{x} & \multicolumn{2}{c|}{y} & \multicolumn{2}{c|}{Start Time} & \multicolumn{2}{c|}{Stay Duration} & \multicolumn{2}{c}{POI} \\
\midrule
      &      & AU $\times 10^{-3}$ & EU $\times 10^{-5}$ & AU $\times 10^{-3}$ & EU $\times 10^{-5}$ & AU $\times 10^{-3}$ & EU $\times 10^{-5}$ & AU $\times 10^{-3}$ & EU $\times 10^{-5}$ & AU $\times 10^{-3}$ & EU \\
\midrule



1-week & 2-week & 1.571 & 1.313 & 2.465 & 2.088 & 0.656 & 7.578 & 0.165 & 0.274 & 34.677 & 1.180 \\
2-week & 2-week & 1.919 & 1.190 & 3.130 & 1.787 & 0.916 & 4.975 & 0.137 & 0.170 & 31.265 & 1.145 \\ 
3-week & 2-week & 1.344 & 0.707 & 2.442 & 1.290 & 0.461 & 0.996 & 0.128 & 0.072 & 44.665 & 1.098 \\

\bottomrule
\end{tabular}%
}
\label{appendix:tab:uc_train_data_relation}
\end{table*}